\documentclass[review]{elsarticle}

\usepackage{lineno,hyperref}

\usepackage{color}
\usepackage{algorithm}
\usepackage{multirow}
\usepackage{hhline}
\usepackage[noend]{algpseudocode}
\usepackage{amsmath}
\usepackage{amsfonts}       
\usepackage{nicefrac}       
\usepackage{algorithm}
\usepackage{hhline}
\usepackage[normalem]{ulem}
\usepackage[font=scriptsize]{subcaption}

\newtheorem{thm}{Theorem}
\newtheorem{lem}{Lemma}
\newdefinition{rmk}{Remark}
\newdefinition{dfn}{Definition}
\newproof{pf}{Proof}
\newproof{pot}{Proof of Theorem \ref{thm2}}

\modulolinenumbers[5]

\journal{Journal of Information Sciences}

\begin{document}

\begin{frontmatter}

\title{Towards a More Reliable Privacy-preserving Recommender System}

\author[mymainaddress,mysecondaryaddress]{Jia-Yun Jiang\corref{mycorrespondingauthor}}
\author[mythirdaddress]{Cheng-Te Li}
\author[mymainaddress]{Shou-De Lin}



\address[mymainaddress]{Department of Computer Science and Information Engineering, National Taiwan University, Taiwan}
\address[mysecondaryaddress]{Institute of Information Science, Academia Sinica, Taiwan}
\address[mythirdaddress]{Department of Statistics, National Cheng Kung University, Taiwan}

\begin{abstract}
This paper proposes a privacy-preserving distributed recommendation framework, \textit{Secure Distributed Collaborative Filtering} (SDCF), to preserve the privacy of value, model and existence altogether. That says, not only the ratings from the users to the items, but also the existence of the ratings as well as the learned recommendation model are kept private in our framework. Our solution relies on a distributed client-server architecture and a two-stage Randomized Response algorithm, along with an implementation on the popular recommendation model, Matrix Factorization (MF). We further prove SDCF to meet the guarantee of \textit{Differential Privacy} so that clients are allowed to specify arbitrary privacy levels. Experiments conducted on numerical rating prediction and one-class rating action prediction exhibit that SDCF does not sacrifice too much accuracy for privacy.
\end{abstract}


\end{frontmatter}

\section{Introduction}

Collaborative filtering (CF) is one of the most popular models for recommending items
\cite{ricci2011introduction}. Its basic idea is to make recommendation based on the similarity between users or between items.
CF-based models are trained on users' feedbacks of items, which can be divided into two categories: numerical and one-class. Numerical feedback consists of numeric value (e.g., ratings between 1 and 5). One-class feedback is a record for a specific action (e.g., purchase or not). As CF-based models are trained from large data, service providers must collect numerous feedbacks. However, if the servers are untrusted or contain vulnerabilities, the collection of user feedbacks can be a privacy liability due to the data leakage. Even if the servers are curious-but-honest, which means the services are functioning normally, 
feedback data leakage can still cause private attributes and even real identity of users, to be inferred by hackers\cite{calandrino2011you,dwork2006calibrating}.
Data leakage in CF-based recommender can be divided into three categories. The first is \textit{Value Leakage}: exposure of values of feedbacks, such as the rating scores. The second is \textit{Existence Leakage}: exposure of the existence of feedbacks, such as whether a user rates an item. For example, if the attackers know that user $u$ rates a book $i$, they can be quite certain that $u$ has read $i$.
The third is \textit{Model Leakage}: exposure of the trained model. Models are important since given the CF model, attackers can estimate the ratings from any user to any item. 


\begin{figure}
  \centering
  \includegraphics[width=0.8\linewidth]{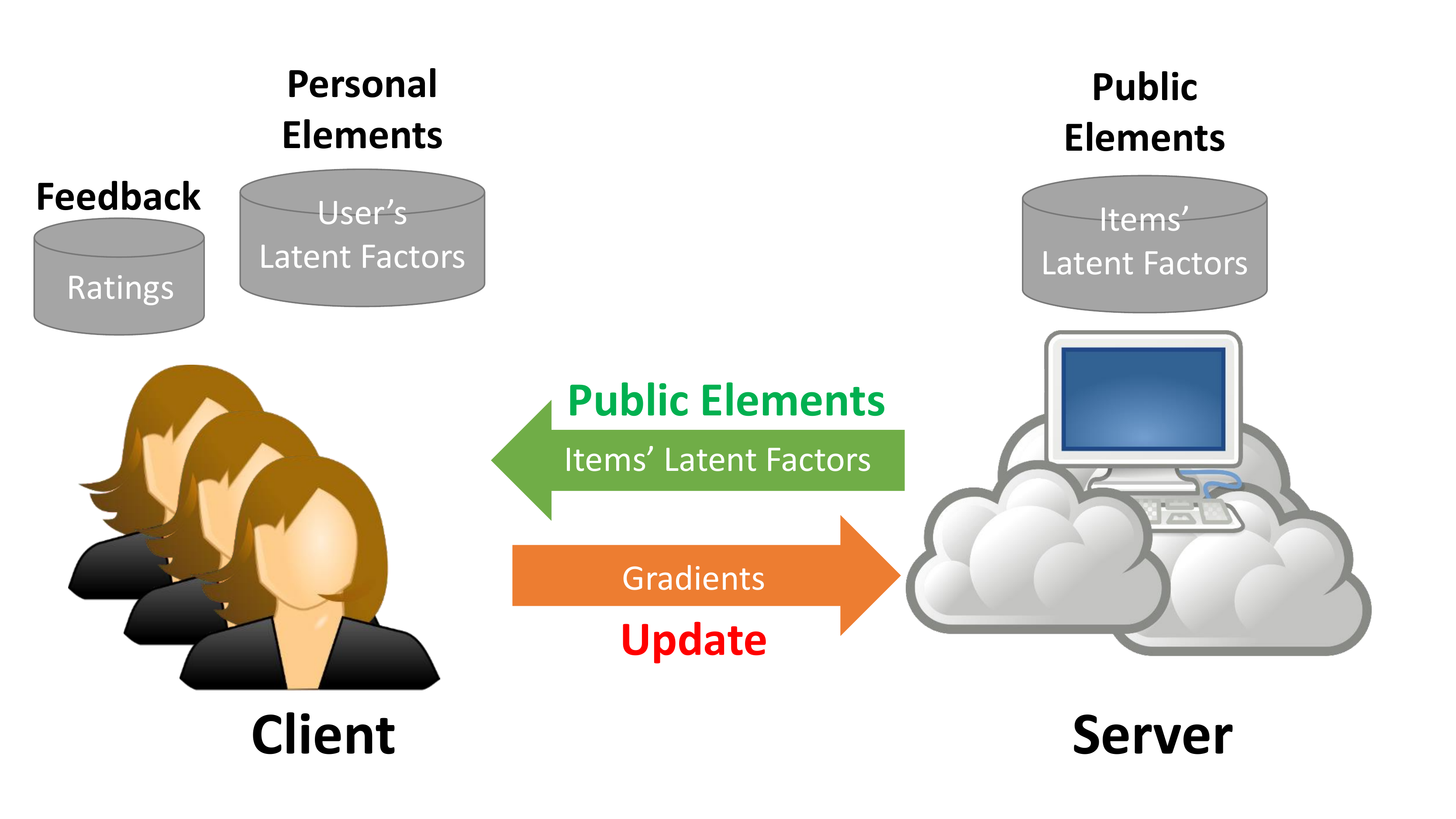}
    \caption{Secured distributed architecture using MF as an example.}
    \label{fig:personal_public}
\end{figure}

In this paper, we propose a client-server framework, \textit{Secure Distributed Collaborative Filtering} (SDCF), to preserve user privacy in CF-based recommender.
We elaborate the idea of SDCF in Fig. \ref{fig:personal_public} that uses Matrix Factorization (MF) as an example. First, we separate the \textit{model} into two disjoint parts: personal elements and public elements. Personal elements are user-specific
while public elements are those common to every user, such as the items' information. We propose to store only public elements on the server, while users keep both their personal ratings and personal elements at client side, to avoid \textit{value} and \textit{model} leakage.
Apparently not all CF-based models meet this requirement. As will be described later, MF satisfies such condition.
The next challenge is how to train a model on such structure since both parts of elements need to be updated frequently while training. We propose that during the training, users download public elements, based on which each user can locally update personal elements, and then send required information to the server to update public elements. Such distributed mechanism naturally prevents both \textit{value} and \textit{model} leakages, because users do not send the ratings to the server, and with only public elements, the server cannot perform recommendation solely. 
To further prevent the \textit{existence} leakage,
we propose a two-stage \textit{Randomized Response algorithm} in SDCF. A strength of SDCF is the guarantee of \textit{Differential Privacy} (DP) \cite{dwork2006calibrating,dwork2014algorithmic}, which is a privacy quantification used in various domains.
With DP, SDCF allows users to adjust their privacy levels 
based on their privacy budgets. 

Previous studies of privacy-preserving recommender systems use various approaches. For example, one of the earliest work \cite{canny2002collaborative} proposes a peer-to-peer network 
that allows personalized recommendation to be generated by other members without disclosing individual user data. \cite{mcsherry2009differentially} first applied DP to CF and is followed by many works including \cite{hua2015differentially,berlioz2015applying,machanavajjhala2011personalized,xin2014controlling,shen2016epicrec,liu2015fast}. Among these studies,  \cite{berlioz2015applying,machanavajjhala2011personalized,liu2015fast,mcsherry2009differentially} require servers to be trusted in some training stages, and \cite{hua2015differentially,nikolaenko2013privacy,shen2016epicrec,canny2002collaborative,xin2014controlling,basu2013privacy} consider the scenario of completely untrusted servers as our work does. Nevertheless, most of them rely on certain conditions or resources. For instance, \cite{shen2016epicrec} requires the category information of items, \cite{xin2014controlling} relies on a small group of users to be public, and \cite{nikolaenko2013privacy,canny2002collaborative} need laborious computation for cryptography algorithms. \cite{hua2015differentially} is probably the most similar to our solution. It assumes nothing other than the ratings, and preserves the privacy of \textit{value} and \textit{model} by a distributed gradient-transmission architecture. Still, it requires a \textit{third-party}, i.e., an additional server between clients and the server of the service provider, as its noise sampling method is vulnerable to \textit{difference attack} defined by \cite{hua2015differentially}. The concern of a third-party is two-fold. First, the third-party may also be compromised, and second the employment of an additional third-party can increase the cost in both building and maintenance. Table \ref{table:related} summarizes the features of different solutions.
To the best of our knowledge, SDCF is the first to preserve \textit{value}, \textit{model}, and \textit{existence} privacies and possess the guarantee of DP. 
More importantly, all feedback data are kept only at client side, i.e., it will not have any chance to be released or transmitted to the untrusted server or any other malicious client.

We choose 
MF \cite{koren2009matrix} to realize SDCF, and develop \textit{Secure Distributed Matrix Factorization} (SDMF), for the following reasons. First, in MF, it is natural to divide the model into personal (user) \textit{U} and public (item) \textit{V} parts (matrices).
Second, if we use gradient descent-based optimization to train the MF model, the updating of personal elements for different users can be done independently and locally at each client, meaning that no personal information will be transmitted to other clients or to the server. Third, the updating of the public element can be done by aggregating the gradients contributed from clients, meaning that only gradients of the model are transmitted from clients to the server. The attackers cannot recover the original ratings nor personal elements even the gradients are intercepted during transmission. 
In this paper, we adopt MF with \textit{Stochastic Gradient Langevin Dynamics} (SGLD) \cite{liu2015fast}. 
As will be discussed later, SGLD (instead of the popular Stochastic Gradient Descent) enjoys the advantage of preventing user latent vectors (i.e., personal elements) from being deciphered. The experiments show that our model can yield reasonably good recommendation results while guaranteeing decent level of privacy. In Section \ref{discuss}, we will further discuss how to use SDCF in \textit{Factorization Machine}-based model \cite{rendle2012factorization}.

\begin{table}
\centering
\caption{Comparison between Different Solutions under Untrusted Server Scenario
}
\label{table:related}
\begin{tabular}{cccccc}
\hline
\textbf{} & \begin{tabular}[c]{@{}c@{}} without Additional \\Information\end{tabular} &\begin{tabular}[c]{@{}c@{}} without \\3rd Party\end{tabular} & \begin{tabular}[c]{@{}c@{}} Existence\\Protected \end{tabular} & \begin{tabular}[c]{@{}c@{}} Model\\Protected \end{tabular} & \begin{tabular}[c]{@{}c@{}} Value\\Protected \end{tabular} \\ \hline
 \cite{canny2002collaborative} & $\surd$ & $\surd$ &  & $\surd$ & $\surd$ \\
\cite{nikolaenko2013privacy} \ & $\surd$ &  & $\surd$ & $\surd$ & $\surd$  \\
\cite{basu2013privacy} \ & $\surd$ & $\surd$ &  &  & $\surd$  \\
\cite{xin2014controlling} &  & $\surd$ &  & $\surd$ & $\surd$ \\
\cite{hua2015differentially} & $\surd$ &  &  & $\surd$ & $\surd$ \\
\cite{shen2016epicrec} &  & $\surd$ & $\surd$ &  & $\surd$ \\ \hline 
\textbf{SDCF} & $\surd$ & \textbf{$\surd$} & \textbf{$\surd$} & \textbf{$\surd$} & \textbf{$\surd$} \\
\hline
\end{tabular}
\end{table}
\section{Preliminaries}

\subsection{Differential Privacy (DP)}
Differential privacy 
provides a mathematical definition to quantify the privacy preserved by an algorithm \cite{dwork2006calibrating,dwork2014algorithmic}.
Here we explain the meaning of DP in recommender systems. If users aim to preserve the \textit{value privacy}, a privacy-preserving algorithm is supposed to lower the server's confidence in identifying the actual value of each rating.

\begin{dfn}[differential privacy]
\label{DP_definition} 
A randomized algorithm $A$ with domain $\mathbb{R}^n$ is $\epsilon$-differentially private, if and only if any two datasets $X, X' \in \mathbb{R}^n$ contain at most one different record (i.e., the hamming distance $d(X,X')=1$), and for possible anonymized output $O \subseteq Range(A)$:
\begin{equation*}
\frac{Pr[A(X)\in O]}{Pr[A(X')\in O]} \leq e^{\epsilon},
\end{equation*}
where the probability $Pr$ is taken over the randomness of algorithm $A$, and $\epsilon$ is positive. Lower values of $\epsilon$ indicates higher degree of privacy guaranteed.
\end{dfn}
With 
Definition 1, we can measure the difference between the outcomes with and without the presence of an element determined by $\epsilon$. 
However, there is a trade-off between user privacy and the utility of an algorithm. Our model is designed to be a good privacy-preserving algorithm that can maintain the performance while increasing the degree of privacy.

\subsection{Matrix Factorization (MF)}
Let $\mathcal{U}$ and $\mathcal{V}$ be a user set and an item set, and each $r_{ij}$ indicates an observed score that a user $i$ rated an item $j$
in the rating matrix $R \in \mathbb{R}^{|\mathcal{U}|\times|\mathcal{V}|}$.
MF aims to make
$UV^{T} \approx R$,
where $U\in \mathbb{R}^{|\mathcal{U}|\times K}$ and $V\in \mathbb{R}^{|\mathcal{V}|\times K}$ for a given dimension number $K$.
The objective is to find $U$ and $V$ such that the 
loss function $L(U,V)$ can be minimized:
\begin{equation}
\label{mf_objective}
\min_{U,V}{L(U,V)}=\min_{U,V}{\sum_{i, j \in R}{(r_{ij}-u_iv_j^T)^2}+\lambda(||U||_2^2+||V||_2^2 ),
}
\end{equation}
where $\lambda$ is a given regularization factor to prevent overfitting. 
Stochastic Gradient Descent (SGD) is a popular technique to minimize the objective function of MF.
It aims to iteratively learn $u_i$ and $v_j$ according to the following updating rules:
\begin{equation}
\label{sgd_update}
\begin{aligned}
\hat{\nabla}_{u_i}L(u_i, v_j) =-e_{ij}v_j+\lambda u_i ,\\ 
\hat{\nabla}_{v_j}L(u_i, v_j) =-e_{ij}u_i+\lambda v_j ,\\ 
u_i \leftarrow u_i-\eta \hat{\nabla}_{u_i}L(u_i, v_j),\\
v_j \leftarrow v_j-\eta \hat{\nabla}_{v_j}L(u_i, v_j),\\
\end{aligned}
\end{equation}
where $e_{ij} =r_{ij}-{u_i}{v_j}^{T}$ is the error of the prediction and $\eta$ is the learning rate.
These updating rules allow the gradients of $u_i$ and $v_j$ to be computed independently,
and thus enables the deployment
in a distributed environment. However, applying SGD in a distributed manner may expose $u_i$ (i.e., \textit{model} leakage) during gradients transmission. 
To remedy this problem, we introduce SGLD below.

\subsection{Stochastic Gradient Langevin Dynamics (SGLD)}
SGLD \cite{welling2011bayesian} is an optimization method, which combines SGD and Langevin Dynamics that samples from the posterior distribution
to prevent overfitting. Specifically, based on the Bayesian view of MF, we can convert the objective into a maximization of posterior distribution of $U$ and $V$ given the observed ratings and parameters:
\begin{equation}
\label{bayes_sgld}
\begin{aligned}
p(U, V|R,\lambda_r,\Lambda_u,\Lambda_v) \propto p(R|U,V,\lambda_r)p(U|\Lambda_u)p(V|\Lambda_v),
\end{aligned}
\end{equation}
where $\lambda_r$ is the global regularization term, $\Lambda_u$ and $\Lambda_v$ are diagonal matrices generated by Gamma distribution for the regularization of $u_i$ and $v_j$. Since a hypothesis of normal distribution $\mathcal{N}$ can be adopted in Eq. \ref{bayes_sgld}, the log-likelihood can be derived from:
\begin{equation}
\label{equation_F(U,V)}
\begin{aligned}
F(U,V) &= \ln p(U, V|R,\lambda_r,\Lambda_u,\Lambda_v)\\
&=\ln p(R|U,V,\lambda_r)+\ln p(U|\Lambda_u)+ \ln p(V|\Lambda_v) + C\\
&= \ln \mathcal{N}(R|UV^T,\lambda_r^{-1})+ \ln \mathcal{N}(U|0,\Lambda_U^{-1})+ \ln \mathcal{N}(V|0,\Lambda_V^{-1}) + C,
\end{aligned}
\end{equation}
where $C$ is a constant.
To maximize the log-likelihood and also follow normal distribution to add noises into gradients to avoid overfitting, the updating rules of SGLD in the $t$-th iteration become:
\begin{equation*}
\begin{aligned}
\hat{\nabla}_{u_i}F(u_i, v_j) =&-e_{ij}v_j+u_i\Lambda_{u}, \\
\hat{\nabla}_{v_j}F(u_i, v_j)=&-e_{ij}u_i+v_j\Lambda_{v}, \\
\eta_t =&\frac{\eta_0}{t^{\gamma}},\\
u_i \leftarrow &u_i-\eta_t \hat{\nabla}_{u_i}F(u_i, v_j)+\mathcal{N}(0,\eta_t\mathtt{I}),\\
v_j \leftarrow &v_j-\eta_t \hat{\nabla}_{v_j}F(u_i, v_j)+\mathcal{N}(0,\eta_t\mathtt{I}),
\end{aligned}
\end{equation*}
where $\eta_0$ is the initial learning rate, $\gamma$ is the decay factor, and $I$ is the identical matrix of size $K$.
As Liu et al. \cite{liu2015fast} has proved that introducing noises to SGLD can guarantee the differential privacy. As will be shown later on, our model using SGLD can avoid the exposure of \textit{model}.

\section{Methodology}


\subsection{Framework Overview}
\begin{figure}
\centering
\includegraphics[width=1.0\linewidth]{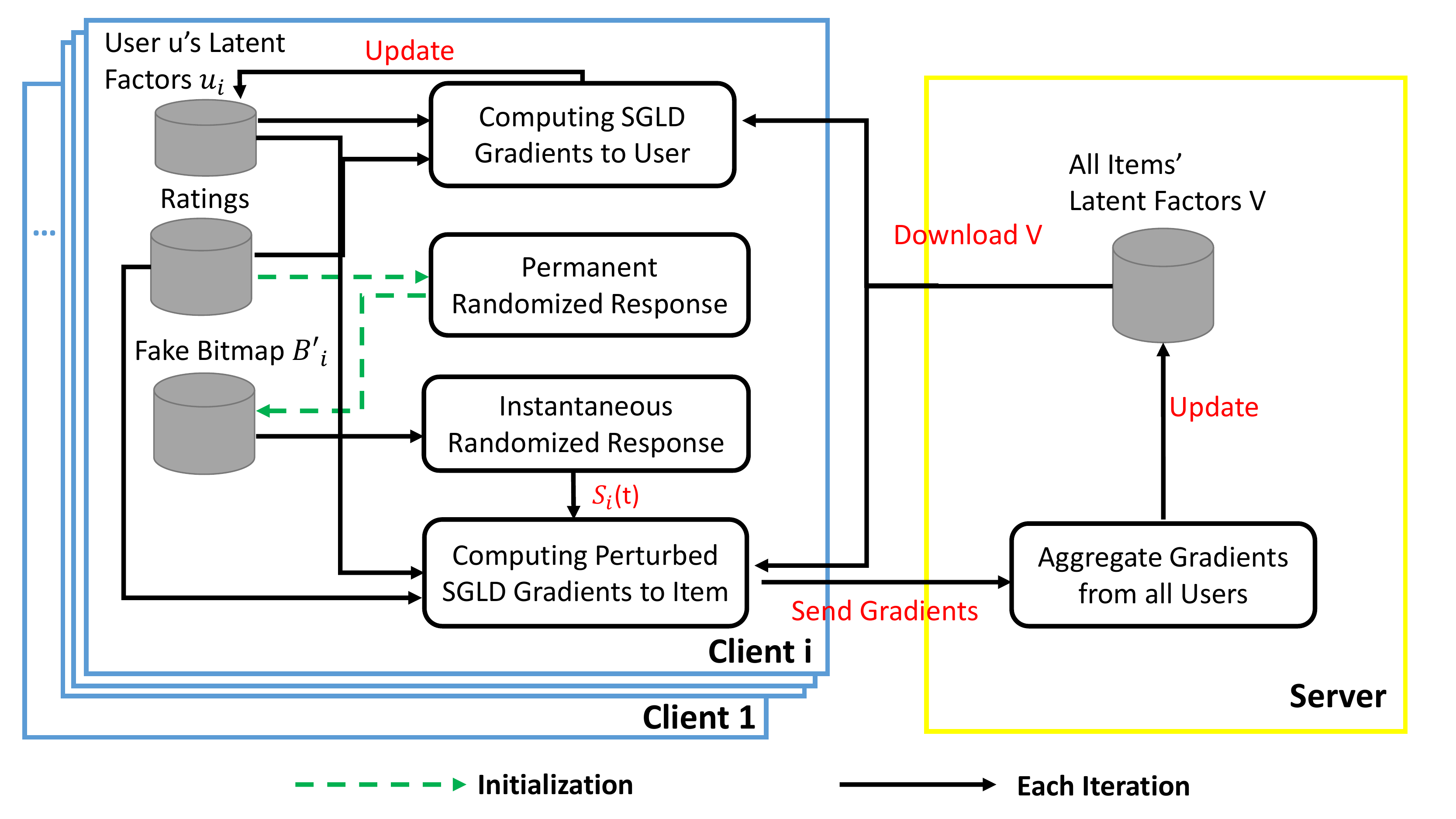}
\caption{The overview of SDMF framework.}
\label{fig:architecture}
\end{figure}

The overview of the SDMF framework is exhibited in Fig. \ref{fig:architecture}. It is a distributed framework consisting of several clients and the server. The key idea is to divide the model into the private (user) parts and a public (item) part, and transmit only gradients to avoid leakage. 
Since MF learns two matrices, items' and users' latent factors, by iteratively performing updates, we consider items' latent factors as the public elements in SDCF, thereby being the only information stored in the server, as illustrated in Fig. \ref{fig:personal_public}. In addition, users' latent factors are treated as personal elements and thus are stored with their ratings at client side only. With SGD to optimize MF's objective (Eq. \ref{mf_objective}), we define the gradients (denoted by $grad(v_j)$ for the latent factors of item $i$ as:
$ grad(v_j)=\eta \hat{\nabla}_{v_j}L(u_i, v_j) $.
The updating rule in Eq. \ref{sgd_update} can be rewritten as:
\begin{equation*}
  \begin{aligned}
	  v_j \leftarrow v_j-grad(v_j) .
  \end{aligned}
\end{equation*}
The gradients are the only things transmitted from clients to the server. Specifically, in each iteration, each client needs to download the newest $V$ from the server, then calculates the gradients for $u_i$ and $v_j$. The gradients of $u_i$ are directly used to update $u_i$ stored at the client side while $grad(v_j)$ is sent to the server, which then updates $v_j$. However, we further look into the formation of $grad(v_j)$ and find that
\begin{equation*}
 grad(v_j)=\eta(e_{ij}u_j-\lambda v_j) .
\end{equation*}
Since $v_j$, $\eta$, and $\lambda$ are public, the server can obtain $e_{ij}u_j$. When $u_j$ converges, the server will be able to approximate each $u_j$ given several $e_{ij}u_j$ for different items within a few iterations, because $u_j$ is the greatest common divisor (GCD) of the values of $e_{ij}u_j$.

To avoid the exposure of $u_i$, we propose to use SGLD instead of SGD to protect the gradients by adding Gaussian noises. The gradients of items can be re-written as:
\begin{equation}
\label{igrad}
\begin{aligned}
grad(v_j) &=\eta_t \hat{\nabla}_{v_j}F(u_i, v_j)-\mathcal{N}(0,\eta_t\mathtt{I}) \\
&= \eta_t(e_{ij}u_i-v_j\Lambda_{v})-\mathcal{N}(0,\eta_t\mathtt{I}).
\end{aligned}
\end{equation}
Aside from DP proved by \cite{liu2015fast}, the noises introduced make it hard to recover $u_i$ from $grad(v_j)$ by GCD. In addition, since the noises in each iteration are sampled with different values, we are immune to the \textit{difference attack} mentioned in \cite{hua2015differentially}. Even if the gradients are exposed, the privacy of \textit{model} and \textit{value} can still be preserved.


\subsection{Randomized Response (RR) Algorithm}
\begin{figure}
\centering
  	\includegraphics[width=0.9\linewidth]{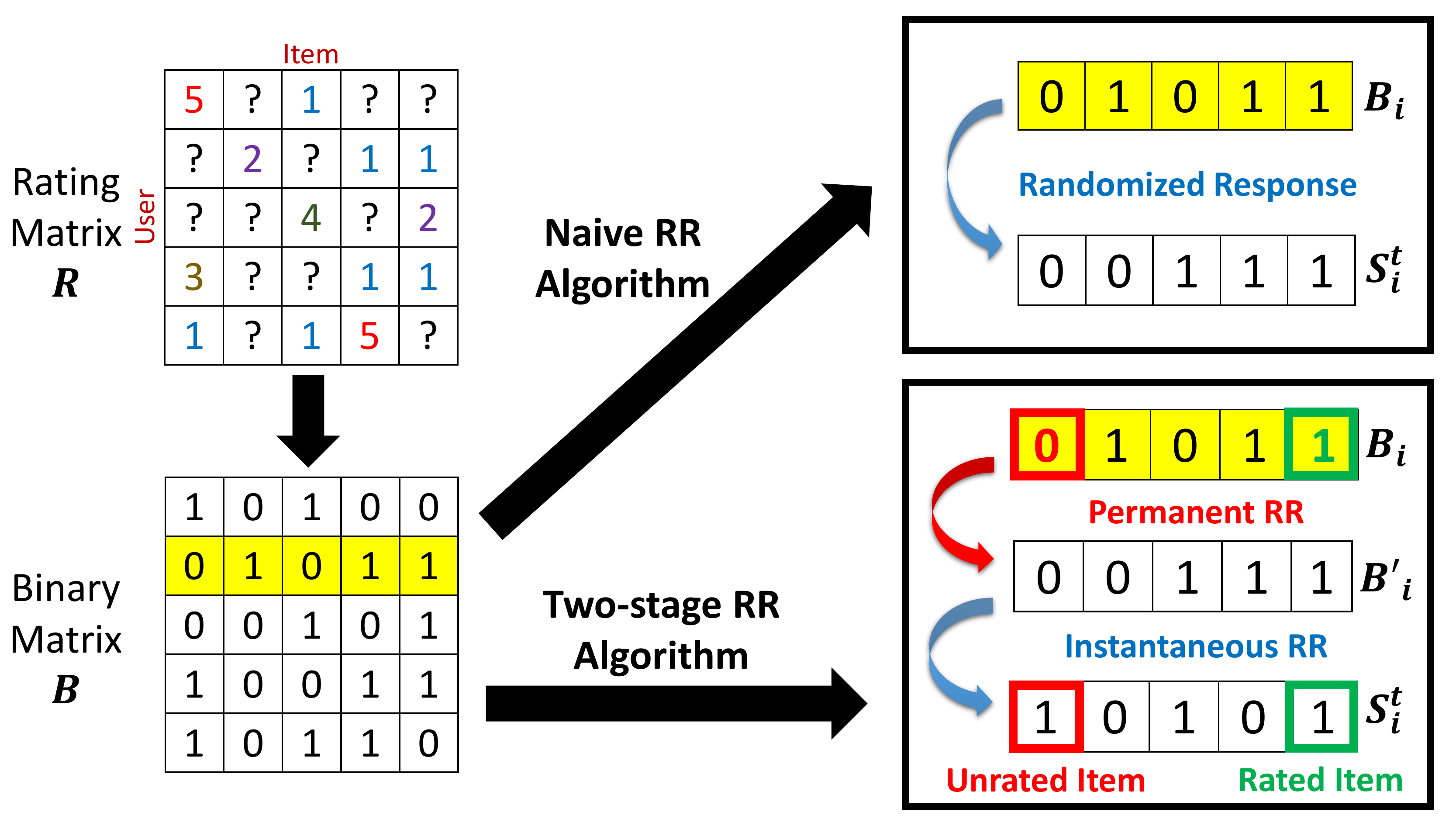}
\caption{Representing the existence of ratings by a binary matrix (left). Using RR to sample a noised bit vector $S^t_i$ from binary vector ${B_i}$ of user $i$ (upper right). Two-stage RR algorithm (lower right). }
    \label{fig:1stage+2stage}
\end{figure}
The \textit{existence} is still exposed due to the action that user $i$ sends $grad(v_j)$ of item $j$ to the server in SGLD. Thus, instead of sending gradients of all rated items, our idea is to send the gradients of some unrated items to the server so that exact items that are rated are unknown. We implement such idea with the technique of randomized response. A straightforward solution consists of two steps. First, we create a binary matrix to represent the existence of ratings, as illustrate in the left part of Fig. \ref{fig:1stage+2stage}, in which each user $i$ maintains a binary vector $B_i$ whose element $B_{ij}$ indicates whether user $i$ rates item $j$. Second, in $t$-th iteration, the client does not send the gradients of all rated items, but the gradients of a perturbed item set determined by a bit array $S^t_{i}$. 
This naive algorithm is shown in the upper right of Fig. \ref{fig:1stage+2stage}. Specifically, the bit array $S^t_{i}$ of user $i$ is used to protect the user's existence by randomizing the rated and unrated items. If $S^t_{ij}=1$, item $j$'s gradient will be sent to the server, otherwise not. We generate the bit array $S^t_{i}$ of item $i$ based on its binary vector $B_i$, given by:
\begin{equation}
\label{IRR_rules}
P(S^t_{ij}=1)=\left\{
    \begin{array}{l}
    p, \text{ if }B_{ij}=0\\
     q, \text{ if }B_{ij}=1\\
    \end{array}
    \right., \forall j \in \mathcal{V},
\end{equation}
where $p$ and $q$ are two given probabilities to determine the number of perturbed unrated and rated items in $S^t_i$. Higher $p$ leads to larger disturbance in the randomization while higher $q$ preserves the actual rating behavior. 
Note that in each iteration $t$, the bit array $S^t_{i}$ of item $i$ is re-generated.

Unfortunately, such simple randomization is vulnerable to the \textit{average attack}. Specifically, untrusted server may collect the gradients for a sufficient number of iterations $T$ and obtain the rating actions by averaging the derived bit array of user $i$ with the following formula:
\begin{equation}
 \frac{1}{T}\sum_{t=1}^{T}(S^t_{ij})\approx \left\{
    \begin{array}{l}
    p, \text{ if }B_{ij}=0\\
     q, \text{ if }B_{ij}=1\\
    \end{array}
    \right..
\end{equation}
To address this, 
we propose to realize the idea of randomized response, which is originally presented by RAPPOR \cite{erlingsson2014rappor}, to construct a secured and distributed recommender system. Note that we are the first that implements the concept of randomized response in distributed recommendation.
The proposed two-stage RR algorithm aims to strengthen the perturbation of rated and unrated items via the binary vector $B_i$ and the bit array $S_{ij}$. 
The \textit{Permanent Randomized Response} (PRR) stage prevents rating exposure caused by the average attack over multiple iterations while the \textit{Instantaneous Randomized Response} (IRR) stage preserves privacy in every single iteration. Note that PRR will be executed only once during initialization and IRR will be executed in the beginning of every iteration.

\subsubsection{Permanent Randomized Response (PRR)}
\label{subsec:prr}
We propose to generate a perturbed binary vector $B'_i$ to permanently replace the original $B_i$ so that we can apply the RR mechanism based on not only the actual rating actions but also some fake ``1''s and ``0''s. With this strategy, the average attacker will not be able to derive $B_i$ since all information sent to server are generated from both fake and actual rating actions. To implement such idea, when initializing the optimization process, we sample the perturbed binary vector $B'_{i}$ for each user $i$ by 
\begin{equation}
\label{PRR_rules}
 B'_{ij}=\left\{
    \begin{array}{l}
    1, \text{with probability }\frac{1}{2}f\\
     0, \text{with probability }\frac{1}{2}f\\
     B_{ij}, \text{with probability }1-f\\
    \end{array}
    \right., \forall j \in \mathcal{V},
\end{equation}
where the parameter $f$ is a probability specified by the client to control the strength of perturbation, and thus determines the percentage of actual rating actions in $B'_i$. With a higher $f$, it will be more private with more perturbation but provide less information for the prediction model. Then $B'_i$ is fixed and become the input to IRR in every iteration. Note that instead of letting the client to provide the probability parameter $f$, the value of $f$ is automatically determined by the differential privacy parameter
$\epsilon_I$, which will be introduced and discussed in Sec. \ref{control_dp}.
 
\subsubsection{Instantaneous Randomized Response (IRR)}
With the perturbed binary vector $B'_i$ derived from PRR, IRR is to implement the naive approach by generating the bit array $S^t_i$. In detail, for each iteration $t$, we sample a bit array $S^t_i$ for every user $i$ based on two probabilities $p$ and $q$ using Eq. \ref{IRR_rules}, in which the original binary vector $B_i$ is replaced by the perturbed version $B'_i$.
Then the client sends gradients of user $i$ to every item $j$ if $S^t_{ij}=1$. Note that
, since the bit array $S^t_i$ is generated by the perturbed binary vector $B'_i$ in Eq. \ref{PRR_rules}, average attack at the server can at most recover the perturbed version $B'_i$. We discuss 
the properties of DP for both PRR and IRR in Sec. \ref{control_dp}.

\subsubsection{Number of Gradients Sent to the Server}
Given we can determine the degree of privacy by adjusting the probability parameters $f$, $p$ and $q$, in this subsection, we aim to discuss how these parameters affect the number of gradients sent to the server which influences the transmission cost of a distributed recommender system. Recall in Eq. \ref{IRR_rules} that the probabilities $p$ and $q$ determine the perturbed rated and unrated items via the bit array. We can estimate the number of gradients sent by first calculating the probabilities of sending the gradients of unrated and rated items, denoted by $p^\star$ and $q^\star$, respectively. 

\begin{lem}
\label{pq_star}
\textit{
The probability of sending gradients of \uline{unrated} item $j$ to the server is:
\begin{equation}
\label{equation_p_star}
\begin{aligned}
p^\star= &P(S^t_{ij}=1|B_{ij}=0) \\
 =&P(B'_{ij}=1|B_{ij}=0)P(S^t_{ij}=1|B'_{ij}=1) \\
 &+P(B'_{ij}=0|B_{ij}=0)P(S^t_{ij}=1|B'_{ij}=0)\\
 =&\frac{1}{2}f\times q+\left(\frac{1}{2}f+(1-f)\right) \times p.
\end{aligned}
\end{equation}
The probability of sending gradients of \uline{rated} item $j$ to the server is:
\begin{equation}
\label{equation_q_star}
\begin{aligned}
q^\star = &P(S^t_{ij}=1|B_{ij}=1) \\
 =&P(B'_{ij}=1|B_{ij}=1)P(S^t_{ij}=1|B'_{ij}=1) \\
 &+P(B'_{ij}=0|B_{ij}=1)P(S^t_{ij}=1|B'_{ij}=0)\\
 =&\left(\frac{1}{2}f+(1-f)\right)\times q+ \frac{1}{2}f \times p.
\end{aligned}
\end{equation}
}
\end{lem}
Let the number of items rated by user $i$ be $h_i$, the expected number of gradients sent to the server is
\begin{equation}
\label{equation_z}
z = h_i\times q^\star+(\mathcal{|V|}-h_i)\times p^\star. 
\end{equation}
It is obvious that the fewer gradients to send, the smaller the transmission cost will be, but the server will get less information to update the model. Therefore, we can allow users to adjust these parameters according to their needs.
In our experiments, for each user $i$, we set $z$ to be $\frac{|R|}{|\mathcal{U}|}$, which is the average number that each user has rated, 
so that the total amount of gradients that the server receives is the same as the original MF and we can make a fair comparison. Note that with $z$ and Theorem 2
, which allows users to specify a privacy budget $\epsilon_I$, we can obtain $p^{\star}$ and $q^{\star}$.

\begin{figure}
\centering
    \includegraphics[width=0.9\linewidth]{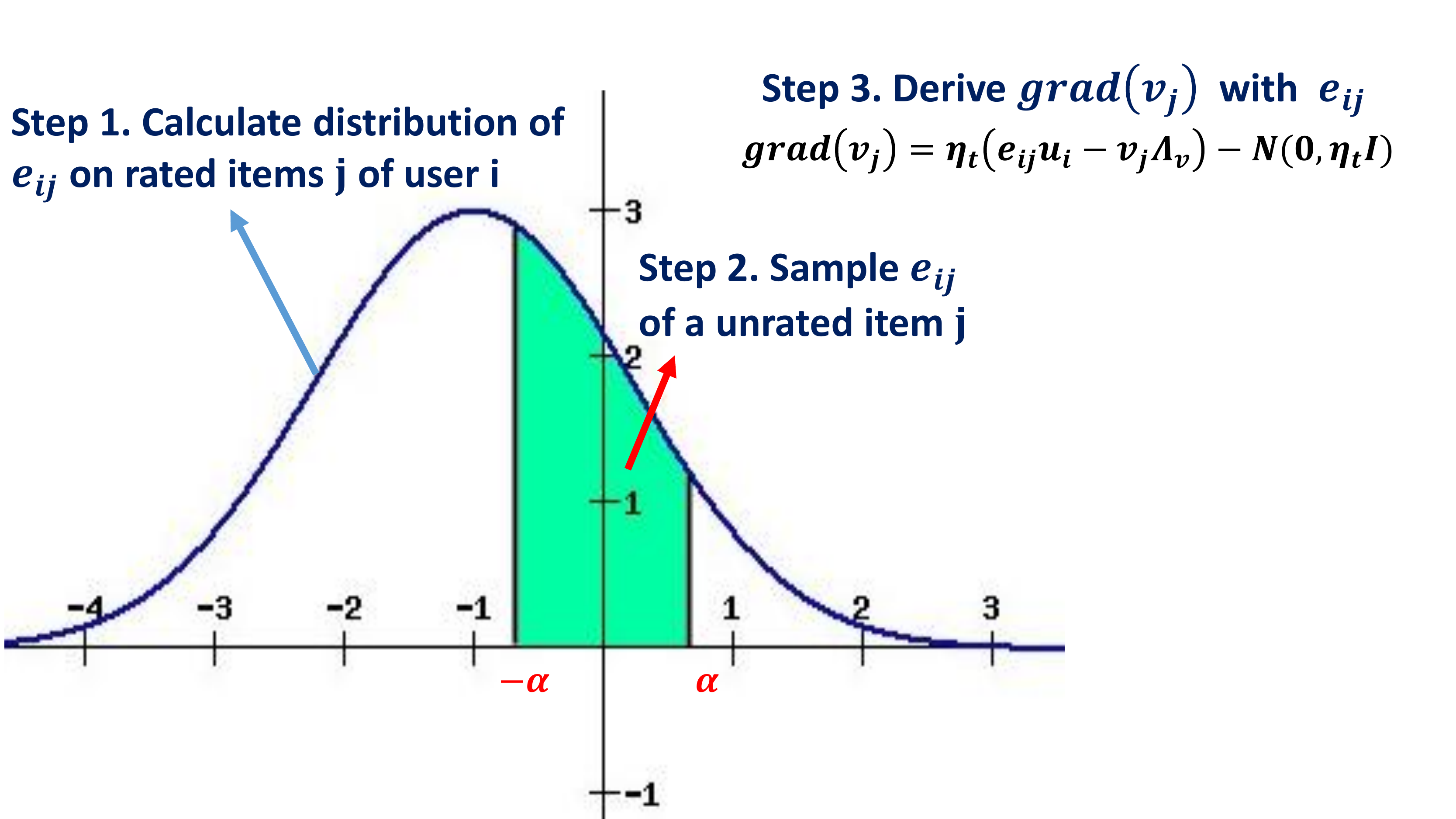}
    \caption{Steps to sample $e_{ij}$ for unrated items.}
    \label{fig:err}
\end{figure}

\subsection{The Gradients of Unrated Items}
\label{sec_unrated}

Now we have only one remaining issue: what to transmit to the server for an item not rated by the user? For a rated item $j$, the user $i$ can simply send its gradient $grad(v_j)=\eta_t(e_{ij}u_i-v_j\Lambda_{v})-\mathcal{N}(0,\eta_t\mathtt{I})$, where $e_{ij}$ is the difference between the predicted and the observed value, $r_{ij}-u_iv_j^T$, as mentioned in Eq. \ref{sgd_update}. However, if item $j$ is unrated, we cannot get $e_{ij}$ because $r_{ij}$ does not exist. Therefore, we propose to sample ``fake'' $e_{ij}$ from the distribution of observed $e_{ij}$ of user $i$. As pointed out in the Bayesian view of SGLD (Eq. \ref{bayes_sgld}), the probabilities of $U$ and $V$ given the observed ratings are assumed to be in normal distributions. Consequently, we can shift $\mathcal{N}(R|UV^T,\lambda_r^{-1})$ in Eq. \ref{equation_F(U,V)} to be zero mean to obtain the distribution of $e_{ij}$, which is also a normal distribution. Therefore, our sampling method for unrated item consists of three steps, as illustrated in Fig. \ref{fig:err}. First, given a certain user $i$, we can count the sample mean $\mu$ and standard deviation $\sigma$ from the rated items so the distribution of $e_{i}$ can be derived. Second, we sample $e_{ij}$ for each unrated item $j$ from $\mathcal{N}(\mu, \sigma)$ within a given range $[-\alpha, \alpha]$. Third, the gradients $grad(v_j)$ with sampled $e_{ij}$ can be derived.

Since $e_{ij}$ is sampled from the distribution of the rated items, the attacker cannot distinguish ``fake'' gradients from actual ones. Besides, the bound in $\alpha$ can constrain the range of $e_{ij}$ to avoid imposing serious noise on the recommendation model. Furthermore, we will discuss how $\alpha$ can be set to control the privacy level in Section \ref{dp_alpha}.


\subsection{Differential Privacy in Our Framework}
\label{control_dp}
An important feature in our framework is that the level of privacy is controllable based on the parameter $f$ in PRR, the probabilities $p$, and $q$ in IRR, as well as the range factor $\alpha$ in computing gradients of unrated items. In this section, we aim to provide theoretical analysis on how the Differential Privacy (DP) is related to these parameters so that one can determine the correspondent parameters for arbitrary privacy levels. Although the total privacy budget scales up with the number of iterations, here we focus on privacy level of each single iteration since the number of iterations that SGLD requires to converge to a certain performance is varied between different datasets. Therefore, we keep it a decision that should be made by the service providers according to their particular goals.

\paragraph{DP of Two-stage Randomized Response Algorithm}
To control the privacy level of the proposed two-stage RR algorithm, we will set a few privacy budgets to control the difficulty for the server to identify the rated items according to the number of times the gradients of items are sent from the user. In the following, we prove the DP in PRR and IRR with the privacy budgets $\epsilon_P$ and $\epsilon_I$, respectively, and derive the relation between these privacy budgets and the probabilities $f$, $p$, and $q$.

\begin{thm}
\label{epsilon_P_therom}
\textit{With a privacy budget $\epsilon_P$, PRR is $\epsilon_P$-differentially private if
\begin{equation*}
\epsilon_P=2h_i\ln\left(\frac{1-\frac{1}{2}f}{\frac{1}{2}f}\right).
\end{equation*}
}
\end{thm}
\begin{pf}
Given the number of rated items $h_i$ and the item set $\mathcal{V}$, for any binary vector $B_i$ with $|\mathcal{V}|$ elements and containing $h_i$ ``1''s, without loss of generality, we can set $B_i$ as $b^{\star} = \{ b_1 = 1, . . . , b_{h_i} = 1, b_{h_i+1} = 0, . . . , b_{|\mathcal{V}|} = 0\}$.

Assume $O_B$ derived from Equation \ref{PRR_rules} in PRR, for all $b'\in O_B$ (for simplicity, we let $b'$ be any possible $B'_i$), given $B_i = b^{\star}$, the probability of $B_i$ being changed into some $B'_i=b'$ is
\begin{equation*}
\begin{aligned}
 P(B'_i=b'|B_i=b^{\star})= &\left(1-\frac{1}{2}f\right)^{b'_1}\left(\frac{1}{2}f\right)^{1-b'_1} \\
& \times \dots \times \left(1-\frac{1}{2}f\right)^{b'_{h_i}}\left(\frac{1}{2}f\right)^{1-b'_{h_i}} \\
& \times \left(\frac{1}{2}f\right)^{b'_{h_i+1}}\left(1-\frac{1}{2}f\right)^{1-b'_{h_i+1}} \\
&\times \dots \times \left(\frac{1}{2}f\right)^{b'_{|\mathcal{V}|}}\left(1-\frac{1}{2}f\right)^{1-b'_{|\mathcal{V}|}} 
\end{aligned}
\end{equation*}
Then for any pair of $B_1$ and $B_2$ in any possible $B_i$, we can derive the ratio in differential privacy (Definition \ref{DP_definition})
 as below:
\begin{equation*}
\begin{aligned}
 \frac{P(B'_i\in O_B|B_i=B_1)}{P(B'_i\in O_B|B_i=B_2)} 
= & \frac{\sum\limits_{b'\in O_B}{P(B'_i= b'|B_i=B_1)}}{\sum\limits_{b'\in O_B}{P(B'_i= b'|B_i=B_2)}} \\
 \leq & \max_{b'\in S}\frac{P(B'_i= b'|B_i=B_1)}{P(B'_i= b'|B_i=B_2)}\\ 
= &  \left(1-\frac{1}{2}f\right)^{2(b'_1+b'_2+\dotsb+b'_{h_i}-b'_{h_i+1}-b'_{h_i+2}-\dotsb-b'_{2h})} \\
& \times \left(\frac{1}{2}f\right)^{2(b'_{h_i+1}+b'_{h_i+2}+\dotsb+b'_{2h_i}-b'_1-b'_2-\dotsb-b'_{h_i})} \\
  \leq &  {\left(\frac{1-\frac{1}{2}f}{\frac{1}{2}f}\right)}^{2h_i}
= e^{\epsilon_P},
\text{when } \epsilon_P = 2h_i\ln\left(\frac{1-\frac{1}{2}f}{\frac{1}{2}f}\right)
\end{aligned}
\end{equation*}
\end{pf}

\begin{thm}
\label{epsilon_I_therom}

\textit{With a privacy budget $\epsilon_I$, IRR is $\epsilon_I$-differentially private, if 
\begin{equation}
\label{equation_epsilon_I}
\epsilon_I=h_i\ln\left(\frac{q^{\star}(1-p^{\star})}{p^{\star}(1-q^{\star})}\right),
\end{equation}
where $p^{\star}$ and $q^{\star}$ follow  
Lemma 1.}
\end{thm}

\begin{pf}
Assume $O_S$ derived from Equation \ref{IRR_rules} in IRR, for $s\in O_S$ (for simplicity, we let $s$ be any possible $S^t_i$), given $B_i = b^{\star}$, the probability of $B_i$ being changed into some $S^t_i=s$ is
\begin{equation*}
\begin{aligned}
P(S^t_i=s|B_i=b^{\star})=(q^\star)^{s_1}(1-q^\star)^{1-s_1}
\times \dots \times (q^\star)^{s'_{h_i}}(1-q^\star)^{1-s'_{h_i}} &\\
\times ({p^\star})^{s'_{h_i+1}}(1-p^\star)^{1-s'_{h_i+1}}
\times \dots \times (p^\star)^{s'_{|\mathcal{V}|}}(1-p^\star)^{1-s'_{|\mathcal{V}|}} & .
\end{aligned}
\end{equation*}
Then for any pair of $B_1$ and $B_2$ in any possible $B_i$, we can derive the ratio in differential privacy (
Definition 1.
) as below:
\begin{equation*}
\begin{aligned}
&\frac{P(S^t_i\in O_S|B_i=B_1)}{P(S^t_i\in O_S|B_i=B_2)} \\ 
=&\frac{\sum\limits_{s\in O_S}P(S^t_i= s|B_i=B_1)}{\sum\limits_{s\in O_S}P(S^t_i= s|B_i=B_2)} 
\leq \max_{s\in O_S}\frac{P(S^t_i = s|B_i=B_1)}{P(S^t_i= s|B_i=B_2)} \\
 \leq & {\left(\frac{q^{\star}(1-p^{\star})}{p^{\star}(1-q^{\star})}\right)}^{h_i} 
= e^{\epsilon_I}, 
\text{when } \epsilon_I =h_i\ln\left(\frac{q^{\star}(1-p^{\star})}{p^{\star}(1-q^{\star})}\right).
\end{aligned}
\end{equation*}
\end{pf}

Practically, $\epsilon_I$ is smaller than $\epsilon_P$ because $\epsilon_P$ is the privacy guarantee of the worst case of suffering from average attack over several iterations. In our implementation, we set $\epsilon_P=2\epsilon_I$ and find $f$ by $\epsilon_P$. Consequently, given $\epsilon_I$ specified by the client, we can obtain $p$ and $q$ by solving Eq. \ref{equation_p_star}, \ref{equation_q_star},\ref{equation_z}, and \ref{equation_epsilon_I}.

\paragraph{DP of Computation of Gradients to Unrated Items}
\label{dp_alpha}
Here we explain how to use the range factor $\alpha$ to control the difficulty for the attacker to distinguish fake gradients of unrated items from gradients of rated items at the server side according to the numerical values of gradients. Note that this privacy level is different from the DP of Randomized  Response algorithm, which is the difficulty of detecting rated items according to discrete results of being sampled or not.

\begin{thm}

\textit{With a privacy budget $\epsilon_g$, gradients for unrated items are $\epsilon_g$-differentially private, if
\begin{equation}
\label{equation_epsilon_g}
\epsilon_g=\ln{\left(\left(\frac{1}{2}erf\left(\frac{x-\mu}{\sigma\sqrt{2}}\right)\arrowvert^{\alpha}_{-\alpha}\right)^{-1}\right)},
\end{equation}
where $\alpha$ is the range factor, and $erf()$ is the Gauss Error Function: 
\begin{equation*}
	erf(x)=\frac{1}{\sqrt[]{\pi}}\int_{-x}^{x}e^{-t^2}dt .
\end{equation*}
}
\end{thm}

\begin{pf}
Based on Section \ref{sec_unrated}, we have the normal distribution of the observed $e_{ij}$: 
\begin{equation*}
f(x) =\frac{1}{\sigma\sqrt{2\pi}}e^{\frac{-(x-\mu)^2}{2\sigma^2}},
\end{equation*}
where $\mu$ and $\sigma$ are mean and standard deviation of the observed $e_{ij}$.
Assume $e'$ is any possible $e_{ij}$, then for any possible item pair of $j_1$ and $j_2$, we derive the ratio in differential privacy (
Definition 1.
) as below:
\begin{equation*}
\begin{aligned}
\frac{P(e_{ij}=e'|j=j_1)}{P(e_{ij}=e'|j=j_2)}
 & \leq \frac{\frac{f(e')}{\int_{-\alpha}^{\alpha}f(x)dx}}{f(e')} 
 =\frac{1}{\int_{-\alpha}^{\alpha}f(x)dx} 
 = \frac{1}{\frac{1}{2}erf\left(\frac{x-\mu}{\sigma\sqrt{2}}\right)\arrowvert^{\alpha}_{-\alpha}}\\ 
 &\leq e^{\epsilon_g},
 \text{when }\epsilon_g=\ln{\left(\left(\frac{1}{2}erf\left(\frac{x-\mu}{\sigma\sqrt{2}}\right)\arrowvert^{\alpha}_{-\alpha}\right)^{-1}\right)}.
\end{aligned}
\end{equation*}
\end{pf}

Since it is difficult to directly solve Eq. \ref{equation_epsilon_g}, in our implementation, we set 
\begin{equation*}
\alpha_{Max}=Max(|\mu+2\sigma|,|\mu-2\sigma|)
\end{equation*}
so that $[-\alpha, \alpha]$ can cover more than 95\% of the ratings in the distribution. Then we do binary search in $(0, \alpha_{Max})$ to find an $\alpha$ that
\begin{equation*}
\begin{aligned}
e^{\epsilon_g-\delta} \leq \left(\frac{1}{2}erf\left(\frac{x-\mu}{\sigma\sqrt{2}}\right)\arrowvert^{\alpha}_{-\alpha}\right)^{-1} \leq e^{\epsilon_g},
\end{aligned}
\end{equation*}
where $\delta$ is a very small number (e.g. $10^{-6}$ in our experiments) so that 
\begin{equation*}
\epsilon_g \approx \ln{\left(\left(\frac{1}{2}erf\left(\frac{x-\mu}{\sigma\sqrt{2}}\right)\arrowvert^{\alpha}_{-\alpha}\right)^{-1}\right)}.
\end{equation*}

\subsection{Implementation Details of SDMF}
We further explain the procedure on the client-side and server-side during the training process of SDMF.

\subsubsection{\textbf{Client-side}}
As step-by-step actions of each iteration $t$ shown in Algorithm \ref{alg:client}. A client first initializes its latent factors, decide the two probabilities used in IRR, $p$ and $q$, and conduct PRR as Equation 8
when $t=0$. In every iteration $t$, it downloads the item latent factors and calculates the prediction error $e_{ij}$ of the rated item. After generating the mean and standard deviation of $e_{ij}$, the client can find a suitable $\alpha$. 
Later, the client will conduct IRR as Equation 
to generate $S^t_i$ of this iteration $t$. For every item $j$ that $S^t_i=1$, if it is a rated item, the client sends the server $j$ and $grad(v_j)$ as in general SGLD; if it is an unrated item, sample $e_{ij}$ in $N(\mu,\sigma)$ (and re-sample if it is not within a bound of $[-\alpha, \alpha]$), sending $grad(v_j)$ calculated with this sampled $e_{ij}$ and $j$ to the server. After the client finishes the sending of $gradients$ to the server, it sends a "finish" signal to the server and updates its user latent factors with the average updates calculated before. 
\begin{algorithm}
	\caption{Client ($i$, $t$)}
    \label{alg:client}
    \begin{algorithmic}[1]
    	\Require Data Input: \{$\mathcal{V}$, $h_i$,$R_i$, $B_i$\}, Model Input: \{$\eta_0$,$\gamma$, $\Lambda_u$, $\Lambda_v$\}, Privacy Input: \{$\epsilon_g$,  $\epsilon_I$, $\epsilon_P$\}
        \If{$t=0$}
        	\State Initialize $u_i$
            \State Find $f$ with $\epsilon_P$
            \State $B'_{i}\leftarrow PRR(B_{i}, f)$
            \State Find $p$ and $q$ with $\epsilon_I$ and $h_c$
        \EndIf
        
        \If{$t>0$}
        	\State $\eta_t\leftarrow \frac{\eta_0}{t^\gamma}$
        	\State Download $V$ from Server
            \State $S_c(t)\leftarrow IRR(B'_{i}, p, q)$
            \State $\bar{u_i} \leftarrow 0$
           	\For{$j \in \mathcal{V}$}
            	\If{$B_{ij}=1$}
                	\State $e_{ij} \leftarrow r_ij-u_iv_j^T$
                    \State $\bar{u_i} \leftarrow \bar{u_i} +(\eta_t(e_{ij}v_j-u_i\Lambda_{u})-\mathcal{N}(0,\eta_t\mathtt{I}))$
                \EndIf
            \EndFor
            \State $\bar{u_i} \leftarrow \frac{\bar{u_i}}{|R_{i}|}$

            \State Calculate $\mu$ and $\sigma$ of $\{e_ij|\forall B_{ij}=1\}$
            \State Find $\alpha$ with  $\mu$, $\sigma$ and $\epsilon_g$
            
            \For{$j \in \mathcal{V}$}
            	\If{$S_i(t)_j=1$ and $B_{ij}=1$}
                	\State $grad(v_j)\leftarrow \eta_t(e_{ij}u_i-v_j\Lambda_{v})-\mathcal{N}(0,\eta_t\mathtt{I})$
                	\State Send ($j$, $grad(v_j)$) to Server
                \EndIf
                \If{$S_i(t)_j=1$ and $B_{ij}=0$}
                	\While{True}
                		\State Sample $e_{ij} \sim N(\mu,\sigma)$
                        \If{$\alpha > x > -\alpha$}
                        	\State break
                        \EndIf
                    \EndWhile
                    \State $grad(v_j)\leftarrow \eta_t(e_{ij}u_i-v_j\Lambda_{v})-\mathcal{N}(0,\eta_t\mathtt{I})$
                	\State Send ($j$, $grad(v_j)$) to Server
                \EndIf                          
            \EndFor
            \State Send "finish" to Server
            \State $u_i \leftarrow u_i-\bar{u_i}$
        \EndIf
    \end{algorithmic}
\end{algorithm}

\subsubsection{\textbf{Server-side}}
Algorithm \ref{alg:server} shows how the server starts to training, obtain $gradients$ and updates items' latent factors. Comparing with Algorithm \ref{alg:client}, it can be seen that the server have to do very little computation. The server is only in charge of controlling the start of a new iteration and updating items' latent factors with whatever $grad(v_j)$ it receives. In this process, nothing but the items' latent factors has to be stored, thereby privacy is preserved. Besides, we can make all clients compute $gradients$ of all their ratings at the same time for each iteration. With this parallelization, the time that the server has to wait for all gradients will be largely reduced, comparing to a centralized training of MF.

\begin{algorithm}
	\caption{Server}
   	\label{alg:server}
    \begin{algorithmic}[1]
    	\Require $\mathcal{U}$, $\mathcal{V}$
        \State Initialize $V$
        \State $t\leftarrow 0$
        \For{$i \in \mathcal{U}$}
            	\State Call Client i to run Client($i$, $t$)
        \EndFor
        
        \State $t\leftarrow 1$
        \While{not converge}
        	\State Initialize $\bar{V}$ with all zero
            \State Initialize $Count=0$
            \For{$i \in \mathcal{U}$}
            	\State Call Client i to run Client(i, t)
         	\EndFor
            \While{True}
            	\State Wait until receiving a $(j, grad(v_j))$
                \State $\bar{V}_j \leftarrow \bar{V}_j+grad(v_j)$
                \State $Count\leftarrow Count+1$
                \If{receive "finish" from all of the Clients}
                	\State break
                \EndIf
            \EndWhile
            \State $V \leftarrow V-\frac{\bar{V}}{Count} $
            \State $t\leftarrow t+1$
        \EndWhile
    \end{algorithmic}
\end{algorithm}

\section{Evaluation}

We conduct experiments to evaluate the utility of SDMF. Specifically, the experiments aim to examine the trade-off between privacy and performance of SDMF with MF. The evaluation is designed to answer three questions: (1) Can SDMF maintain high accuracy while preserving certain level of privacy? (2) How does the privacy budgets $\epsilon_I$ and $\epsilon_g$ in the proposed two-stage RR method affect accuracy? (3) How does SDMF perform on both numerical and one-class rating prediction tasks? Note that the experiments here focus on the effect of using DP to preserve \textit{existence} privacy, so we do not compare with other DP methods such as \cite{hua2015differentially,shen2016epicrec} since their DP algorithms aim to preserve \textit{value} privacy, which SDMF can completely preserve with the architecture.

\subsection{Datasets \& Evaluation Settings}
We use three popular public rating datasets for the experiments: two MovieLens datasets
\footnote{\url{https://grouplens.org/datasets/movielens/}}
(ML-100K and ML-1M) and Netflix data \cite{Bennett07thenetflix}. 
Note that we subsample Netflix data to a subset with 10,000 users and 5,000 items, all with more than 10 ratings to avoid cold-start users, who will contribute little due to the RR algorithm. 
The statistics of MovieLens-100K, MovieLens-1M and the subsampled Netflix dataset are shown in Table \ref{datastat}.
\begin{table}
\centering
\caption{Statistics for the datasets.}
\label{datastat}
\begin{tabular}{crrr}
\hline
Dataset & \multicolumn{1}{c}{\#Users} & \multicolumn{1}{c}{\#Items} & \multicolumn{1}{c}{\#Ratings} \\ \hline 
\textbf{MovieLens-100K} & 943 & 1,682 & 100,000 \\
\textbf{MovieLens-1M} & 6,040 & 3,900 & 1,000,209 \\
\textbf{Netflix (subsampled)} & 10,000 & 5,000 & 573,595 \\ \hline
\end{tabular}
\end{table}

\paragraph{\textbf{Task 1: Numerical Rating Prediction}} 
We randomly split both rating datasets into 80\% as the training set and 20\% as the test set. This random splitting is repeated for 30 times to obtain the average \textit{Root-Mean-Square Error} (RMSE). In order to examine how different privacy levels affect the performance, we vary the controllable parameters, i.e., the privacy budgets $\epsilon_g$ and $\epsilon_I$ in SDMF. As data quality (or noise introduced) is determined by $\epsilon$-differential privacy, higher values of $\epsilon_g$ and $\epsilon_I$ lead to weaker privacy protection, but maintain the accuracy of the recommendation model. Hence, we choose to have a wide range of privacy budgets so that the trade-off between performance and privacy can be observed. The privacy budgets $\epsilon_g$ and $\epsilon_I$ are chosen from $\{4, 1, 0.25, 0.0625\}$, resulting in 16 combinations of privacy budgets in total. Note that we set $\epsilon_P=2\epsilon_I$ so that we do not show the results of different $\epsilon_P$.

The settings of parameters are: $K=50$, $\gamma=0.6$, and , $(\Lambda_u, \Lambda_v) \sim Gamma(1,100)$. The learning rate $\eta_0$ is $5\times 10^{-6}$ for MovieLens-100K, $5\times 10^{-7}$ for MovieLens-1M, and $5\times 10^{-7}$ for Netflix, which are tuned to have better performance based on a validation set made by randomly splitting the training set into 80\%-20\%,
which are tuned to have better performance based on a validation set made by randomly splitting the training set into 80\%-20\%.
Besides, SDMF will be compared to the baselines and competitive methods as listed in the following. 
\begin{itemize}
\item \textbf{Non-private MF:} Performing recommendation using the original MF with SGLD. We compare this non-private version with SDMF to understand how the noises introduced for preserving privacy affect performance.

\item \textbf{Input Perturbation SGLD (ISGLD):}
We make a comparison to a na\"ive strategy of simply adding noises to ratings. Note that this solution can only preserve value, not model nor existence privacies.
We add Laplacian noises to ratings and train MF using SGLD on these perturbed ratings to derive ISGLD. The $\epsilon$-differential privacy of adding Laplacian noises to ratings has been proved in \cite{berlioz2015applying}. 
We compare SDMF with ISGLD with privacy budget $\epsilon$ set to 4 and 2, while the former is the largest number we set for privacy budgets in SDMF and the latter is what reported to have comparable performance as item average baseline by \cite{berlioz2015applying}. 

\item \textbf{SDMF $\alpha = \infty$:} This is an SDMF given no constraint on the sampled fake gradient. Hence, it would lead to the highest privacy level for given fixed $\epsilon_g$ and $\epsilon_I$ in SDMF.

\end{itemize}

\paragraph{\textbf{Task 2: One-Class Rating Action Prediction}.}
To conduct experiments of SDMF on the task of one-class feedback, we choose to predict the rating actions via Bayesian Personalized Ranking Matrix Factorization (BPRMF) \cite{rendle2009bpr} in ML-100K, ML-1M, and the subsampled Netflix dataset. That says, BPRMF is considered MF technique in our SDMF. The split of training and testing is set by the leave-one-out strategy, which is the same as BPRMF\cite{rendle2009bpr}. Specifically, we first randomly select one rating action of each user to be added into the testing set while the training set consists of all rating actions except those in the testing set.
This process is repeated for 10 times to generate the average Area under ROC Curve (AUC), which indicates the correctness of pair-wise ranking between a positive (rated) and a negative (unrated) sample. 
While the gradients of BPRMF is similar to the basic MF, we can accordingly apply BPRMF to SDMF by changing the error $e_{ij}$ in MF to $\frac{-e^{-x_{ijj'}}}{1+e^{-x_{ijj'}}}$ for rated items and $\frac{e^{-x_{ijj'}}}{1+e^{-x_{ijj'}} }$ for unrated items, where $x_{ijj'} = u_iv_j^T-u_iv_{j'}^T$ denotes the distance between the predicted ranking scores of a rated item $j$ and an unrated item $j'$.
The updating rules of using BPRMF in SDMF, termed \textbf{SD-BPRMF}, is shown in 
Lemma 2.
Since the optimization of BPRMF computes gradients for unrated items,
SD-BPRMF does not need the ``fake'' errors, and the privacy budget $\epsilon_g$ is not applied here.

\begin{lem}
\label{lemma_bpr}
\textit{For a rated item $j$ rated by user $i$, SD-BPRMF randomly samples an unrated item $j'$ and updates its latent factors with
\begin{equation*}
\begin{aligned}
	u_i &\leftarrow u_i -\eta_0 \left(\frac{e^{-x_{ijj'}}}{1+e^{-x_{ijj'}} } (-v_j+v_{j'})+u_i\Lambda_{u}\right)+\mathcal{N}(0,\eta_t\mathtt{I}),\\
v_j &\leftarrow v_j -\eta_0 \left(\frac{-e^{-x_{ijj'}}}{1+e^{-x_{ijj'}} } (u_i)+v_j\Lambda_{v}\right)+\mathcal{N}(0,\eta_t\mathtt{I}),\\
v_{j'} &\leftarrow v_{j'} -\eta_0 \left(\frac{e^{-x_{ijj'}}}{1+e^{-x_{ijj'}} } (u_i)+v_{j'}\Lambda_{v}\right)+\mathcal{N}(0,\eta_t\mathtt{I}).
\end{aligned}
\end{equation*}
}
\end{lem}

We compare the performance of SD-BPRMF with the privacy budget $\epsilon_I \in \{4,1,0.25,0.0625\}$ 
and a non-private version of BPRMF using SGLD as the baseline. The settings of parameters are: $K=10$, $(\Lambda_u, \Lambda_v) \sim Gamma(1,100)$, $\gamma=0.6$, and $\epsilon_P = 2\epsilon_I$. 
The learning rate $\eta_0$ is $5\times 10^{-6}$ for both MovieLens datasets (MovieLens-100K and MovieLens-1M), and ${10}^{-7}$ for Netflix dataset, which are chosen according to the validation set. 

\subsection{Experimental Results}
Since the number of training iterations is positively correlated with the transmission overhead in our framework, there is a trade-off between recommendation performance and computational cost while the privacy cost are scaled up by the number of iterations as well. Therefore, we show the experimental results by presenting the learning curve (i.e., RMSE vs. ``number of iterations'' up to 100) for each model. Based on the results, service providers are allowed to choose the most appropriate numbers in their applications.

\begin{figure}
\centering
\begin{subfigure}{1.0\linewidth}
\centering
\includegraphics[width=.35\linewidth]{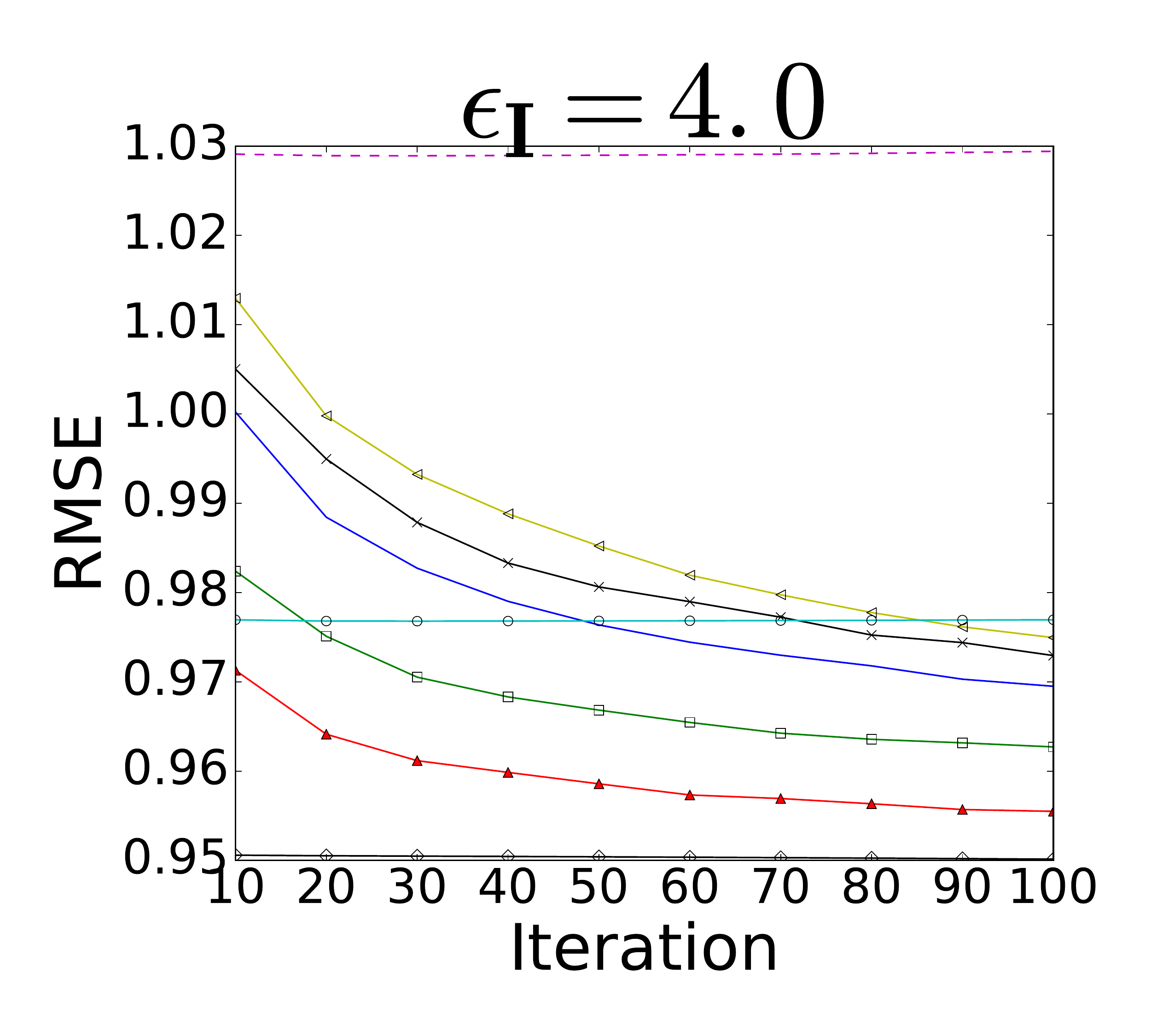}
\includegraphics[width=.35\linewidth]{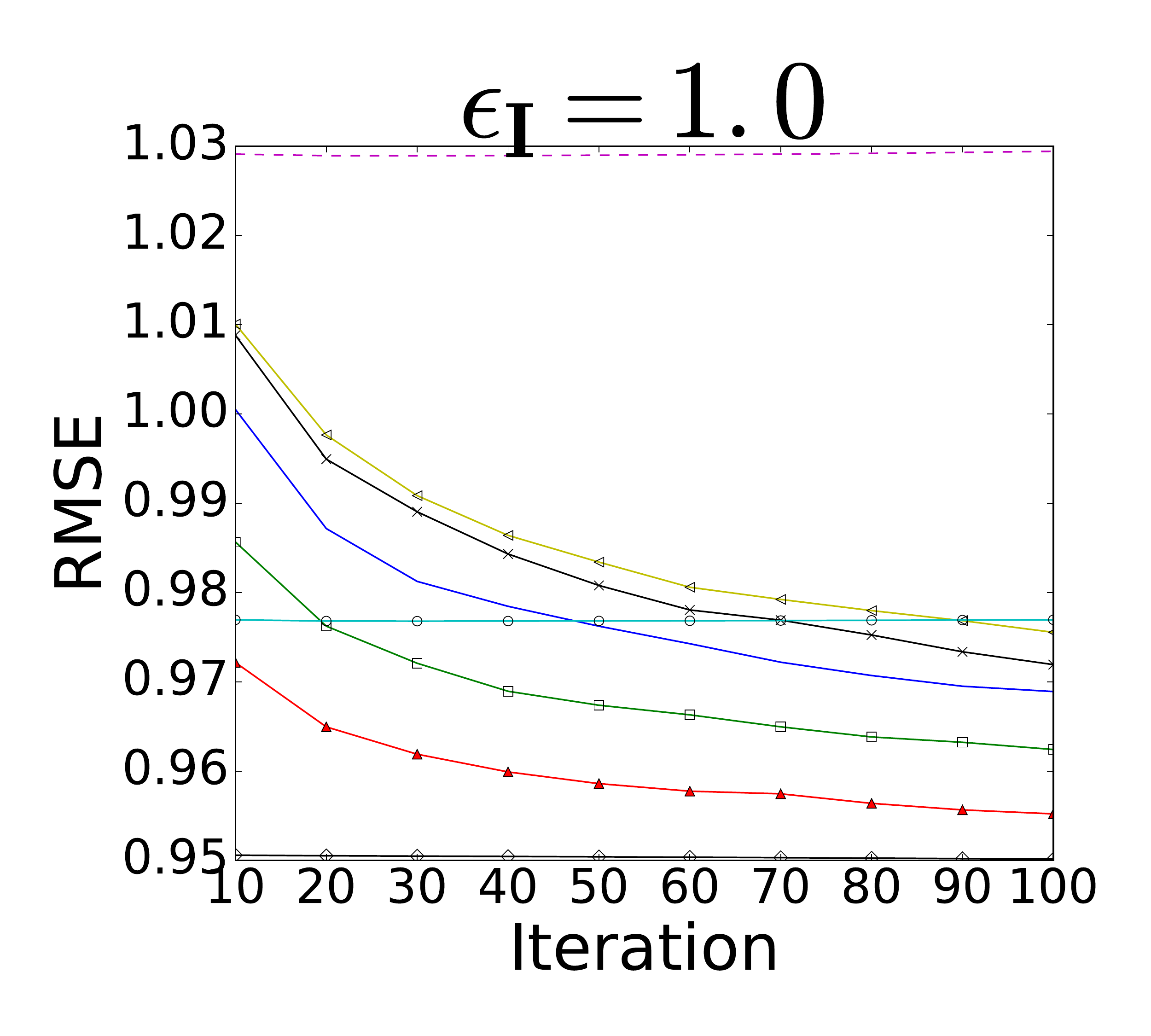}

\includegraphics[width=.35\linewidth]{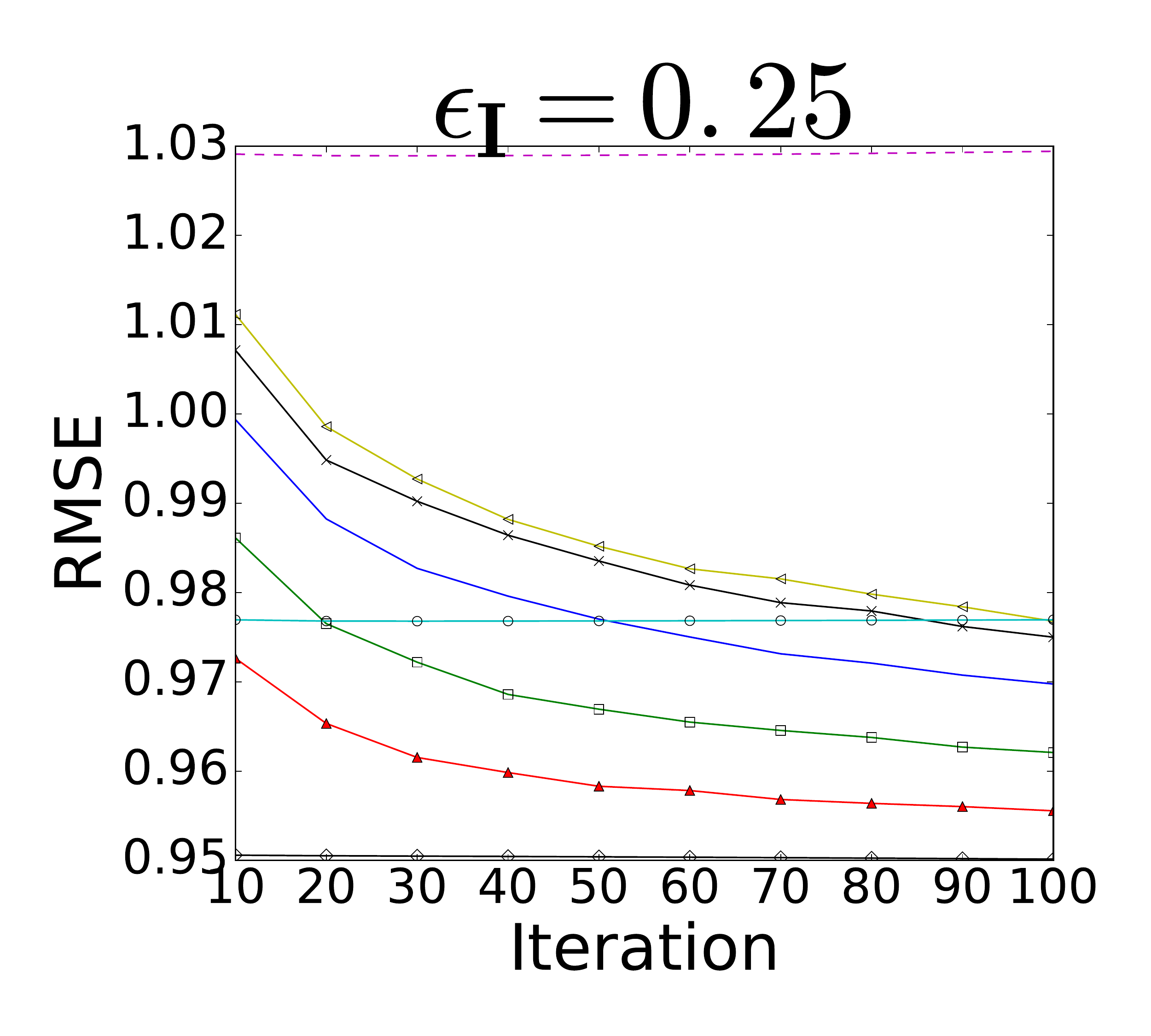}
\includegraphics[width=.35\linewidth]{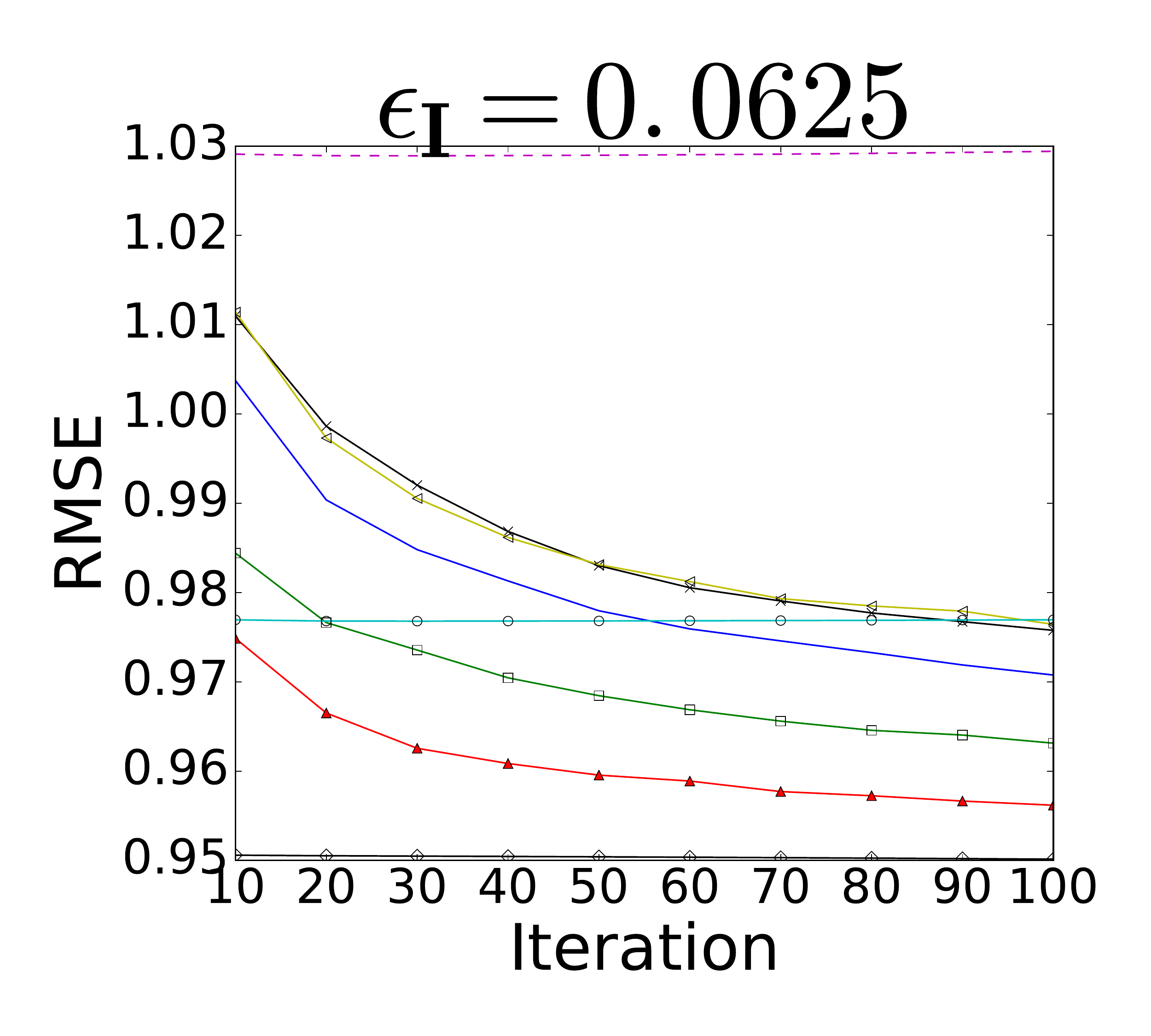}
\includegraphics[width=0.7\linewidth]{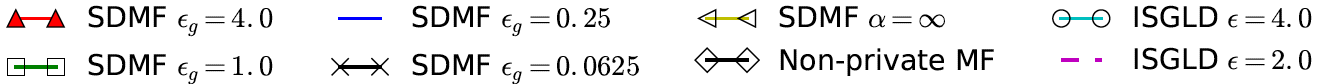}
\caption{Task 1: Comparison of different $\epsilon_g$ with fixed $\epsilon_I$.}
\label{fig:ml100k_curve_g}
\end{subfigure}
\begin{subfigure}{1.0\linewidth}
\centering
\includegraphics[width=.35\linewidth]{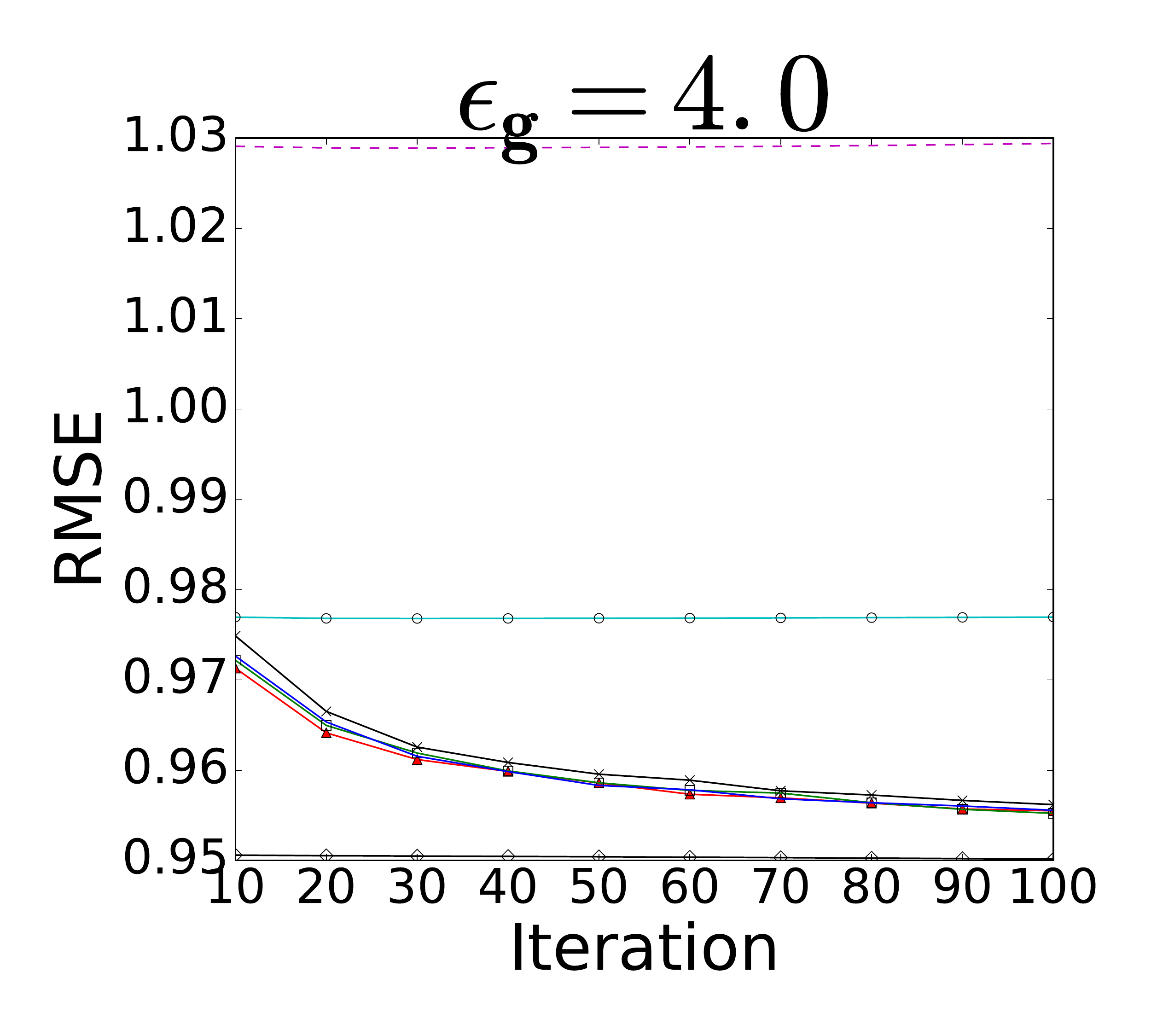}
\includegraphics[width=.35\linewidth]{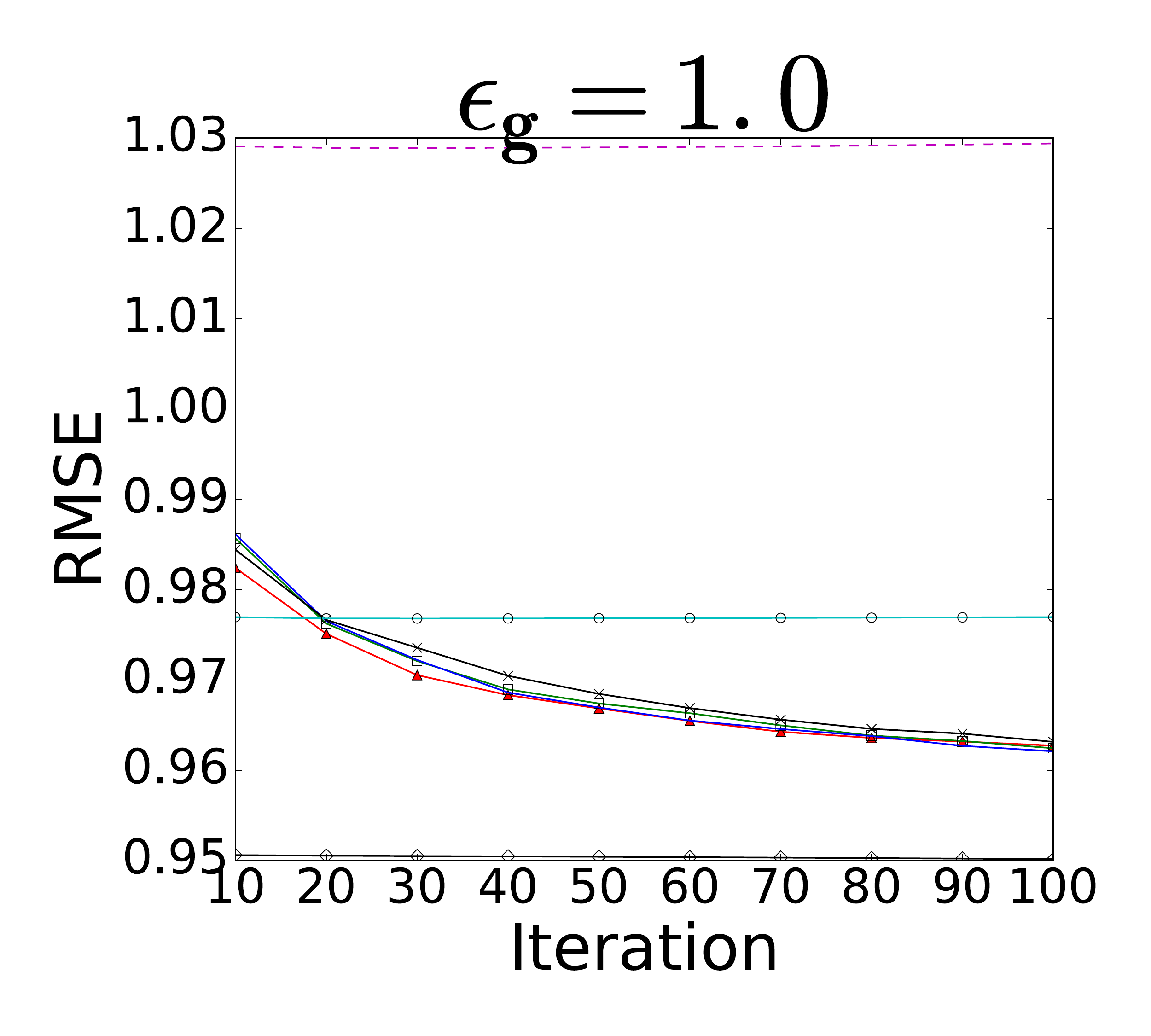}

\includegraphics[width=.35\linewidth]{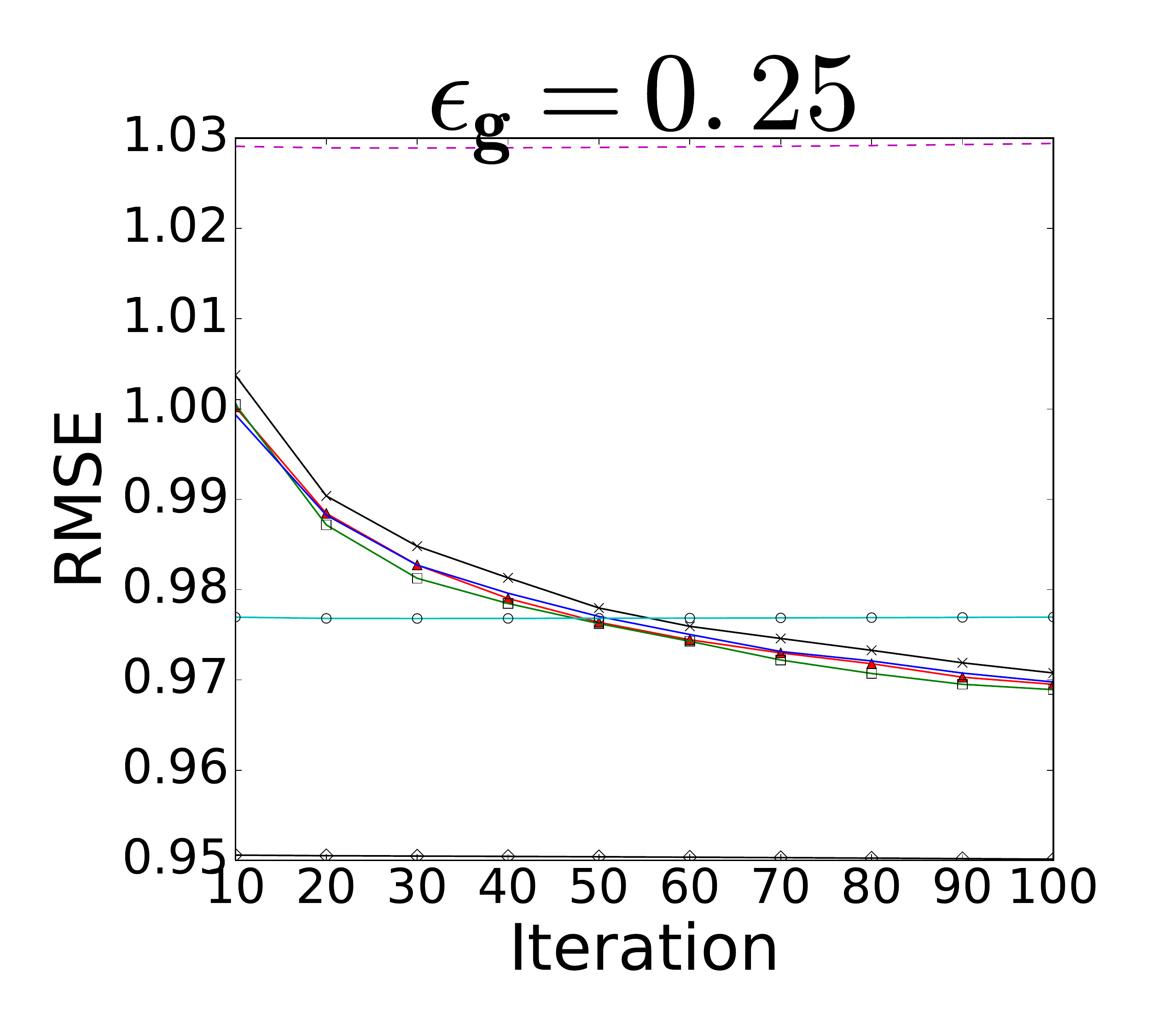}
\includegraphics[width=.35\linewidth]{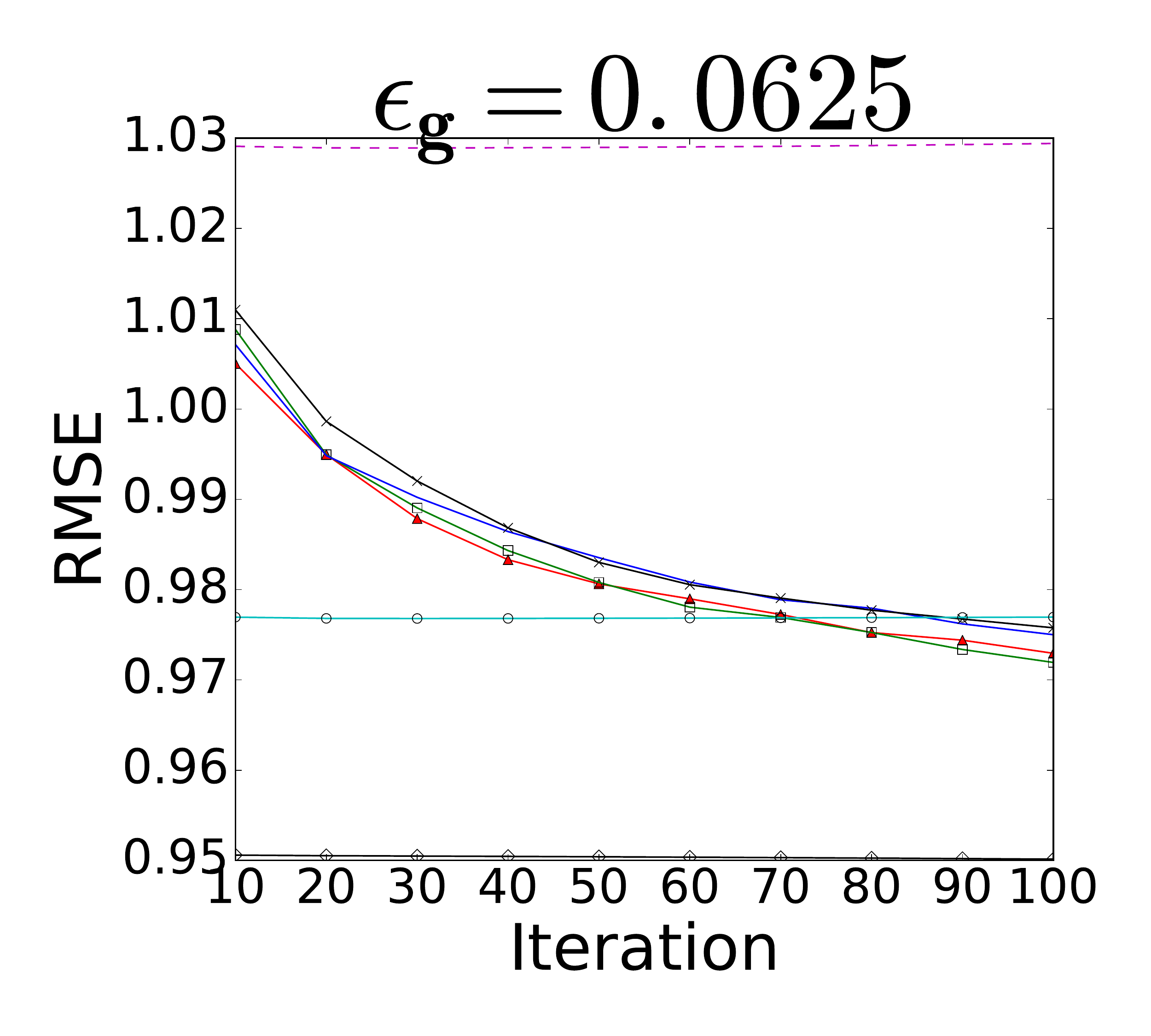}
\includegraphics[width=0.7\linewidth]{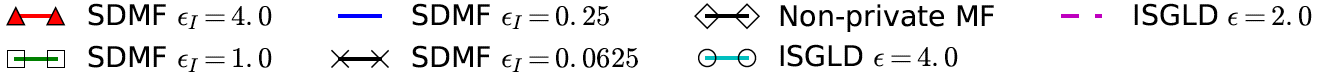}
\caption{Task 1: Comparison of different $\epsilon_I$ with fixed $\epsilon_g$.}
\label{fig:ml100k_curve_I}
\end{subfigure}
\caption{MovieLens-100K}
\end{figure}

\begin{figure}
\begin{subfigure}{1.0\linewidth}
\centering
\includegraphics[width=.35\linewidth]{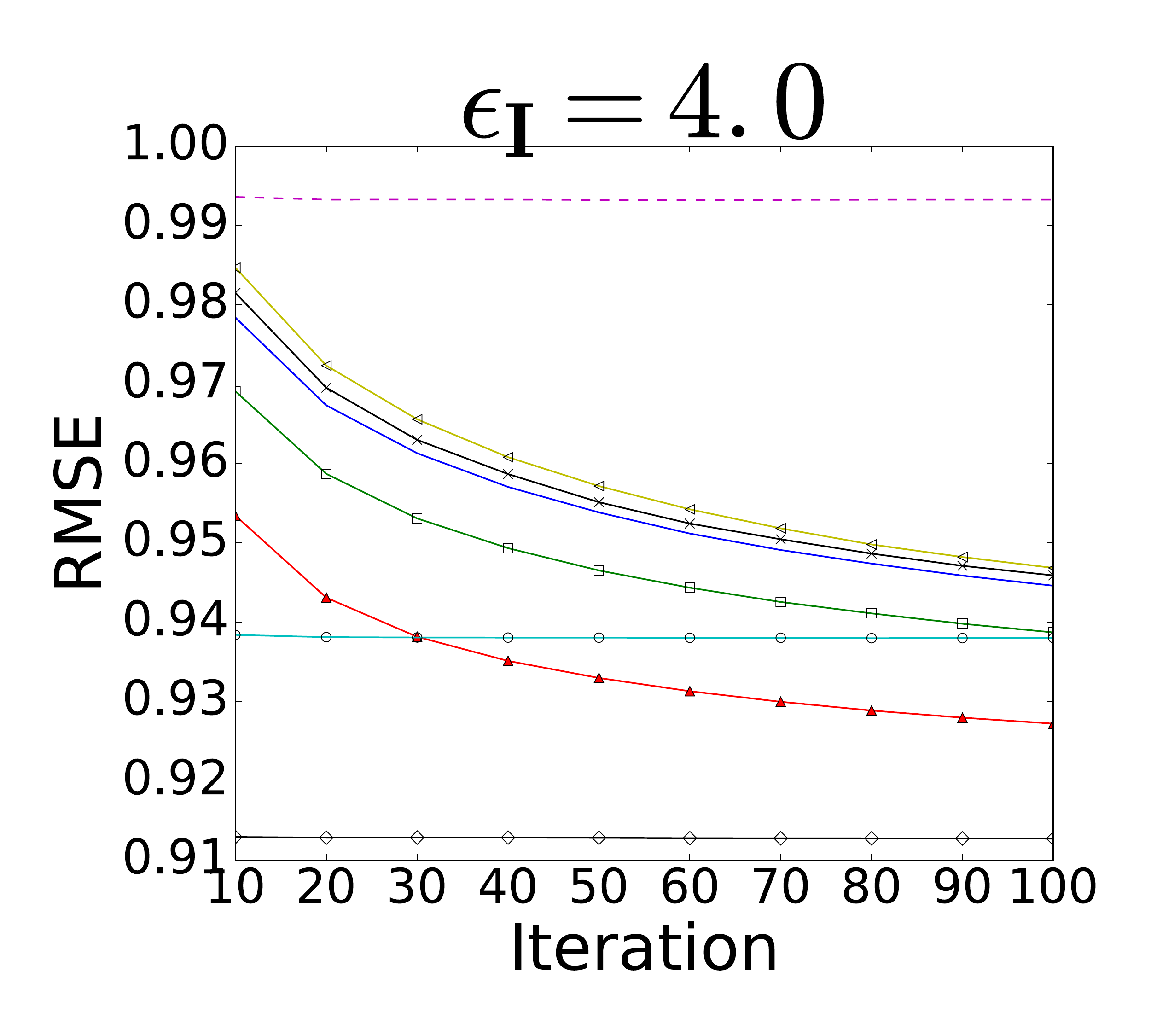}
\includegraphics[width=.35\linewidth]{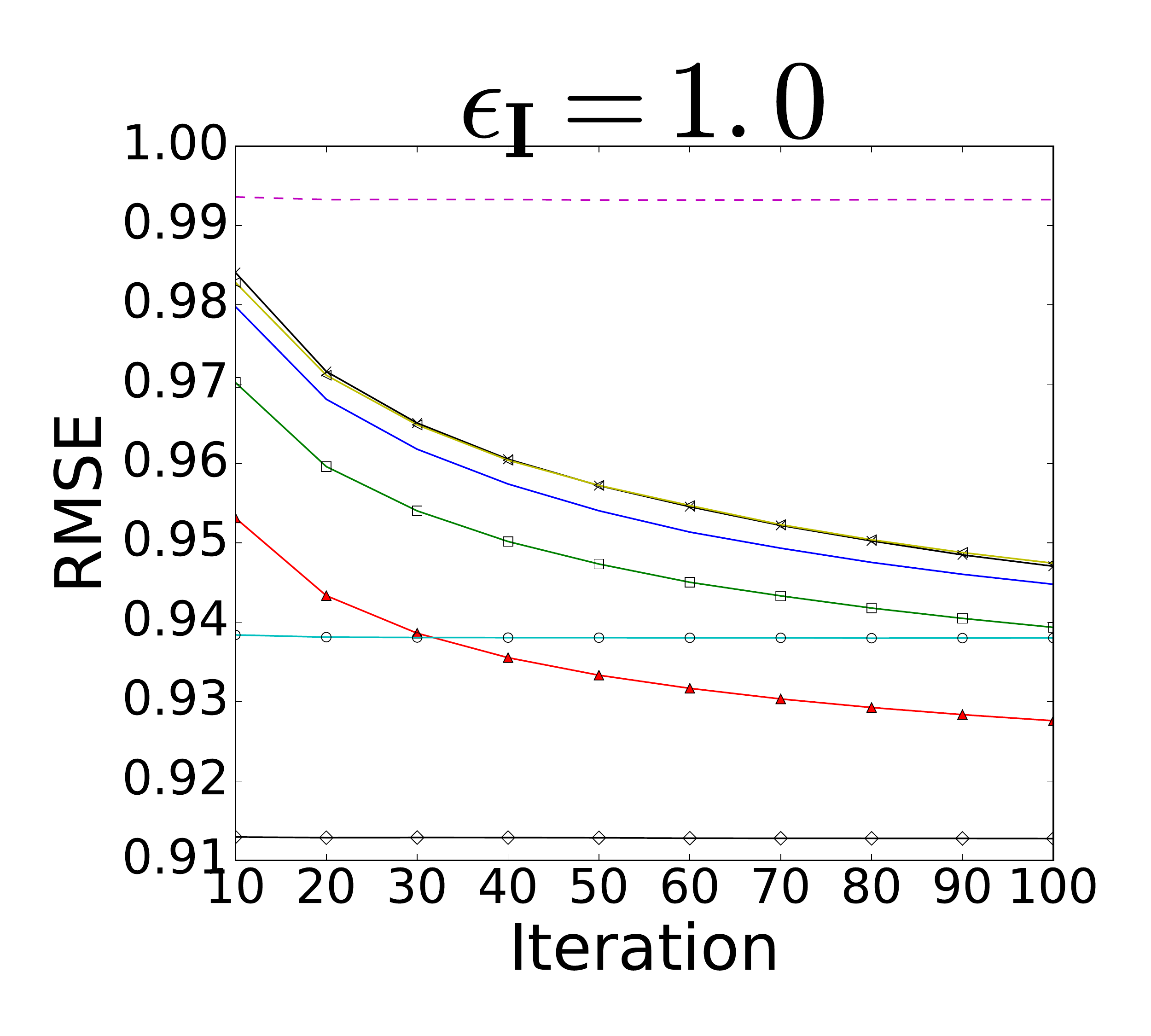}

\includegraphics[width=.35\linewidth]{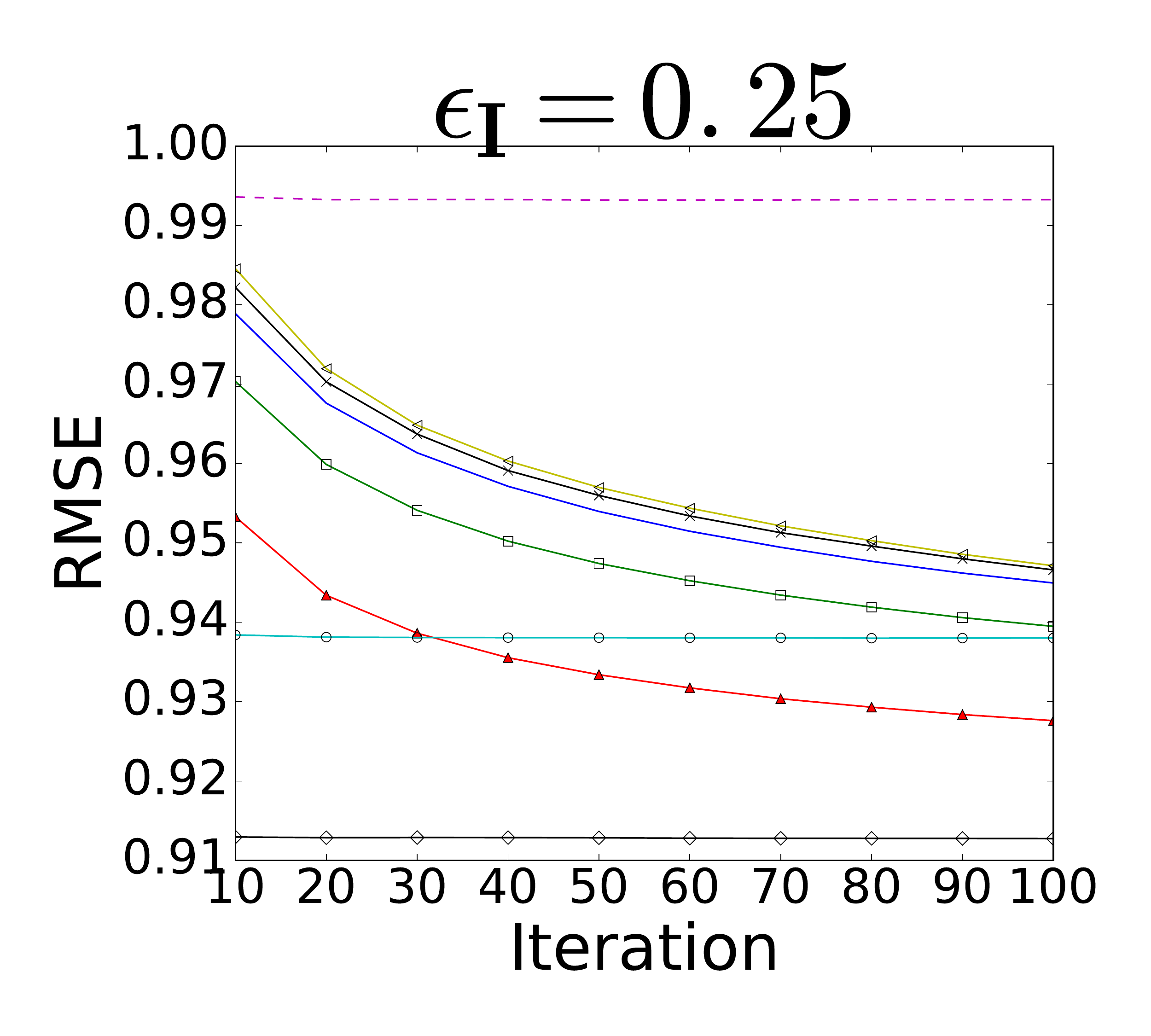}
\includegraphics[width=.35\linewidth]{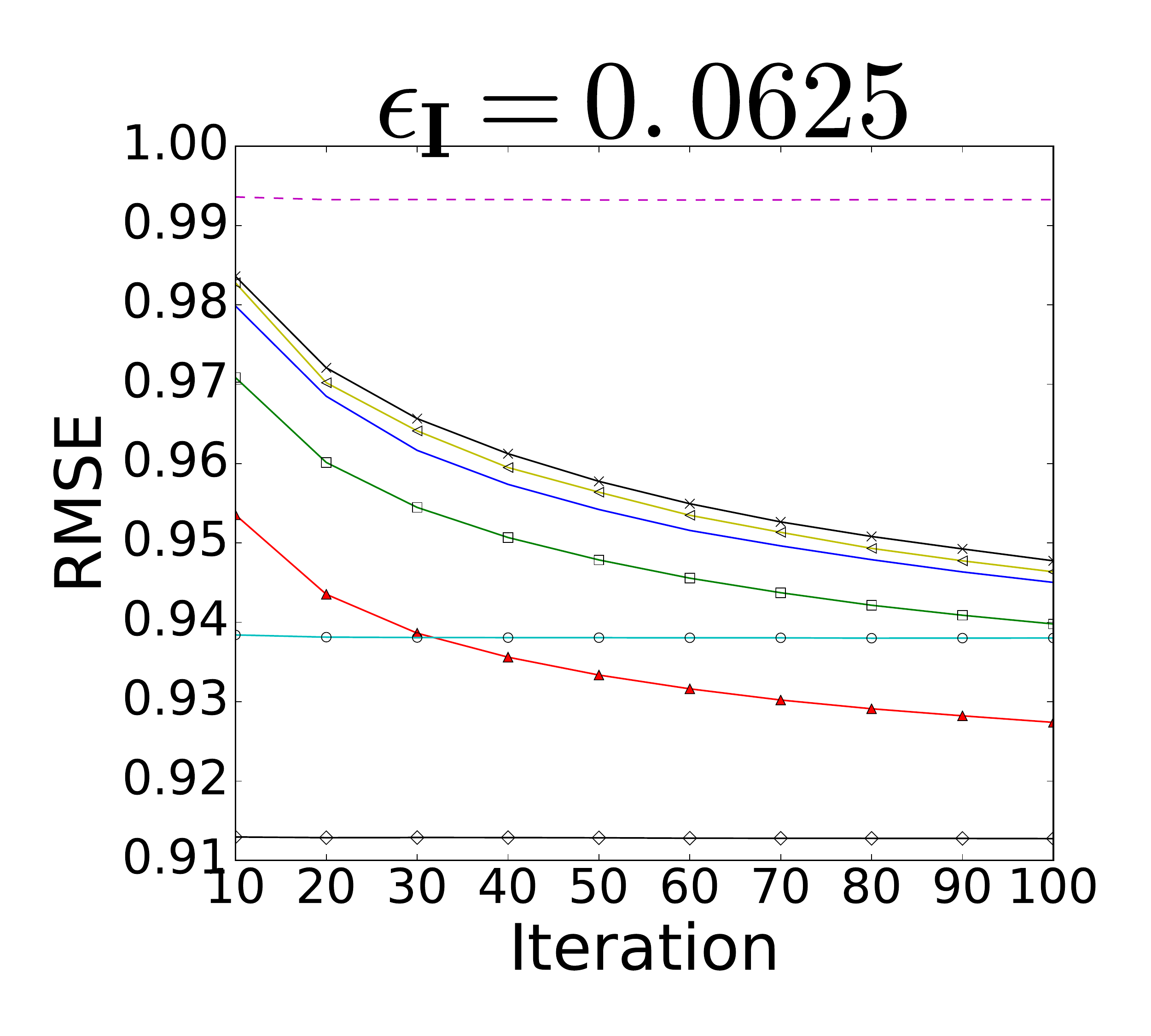}
\includegraphics[width=0.7\linewidth]{legend_epsI.png}
\caption{Task 1: Comparison of different $\epsilon_g$ with fixed $\epsilon_I$.}
\label{fig:ml1M_curve_g}
\end{subfigure}
\begin{subfigure}{1.0\linewidth}
\centering
\includegraphics[width=.35\linewidth]{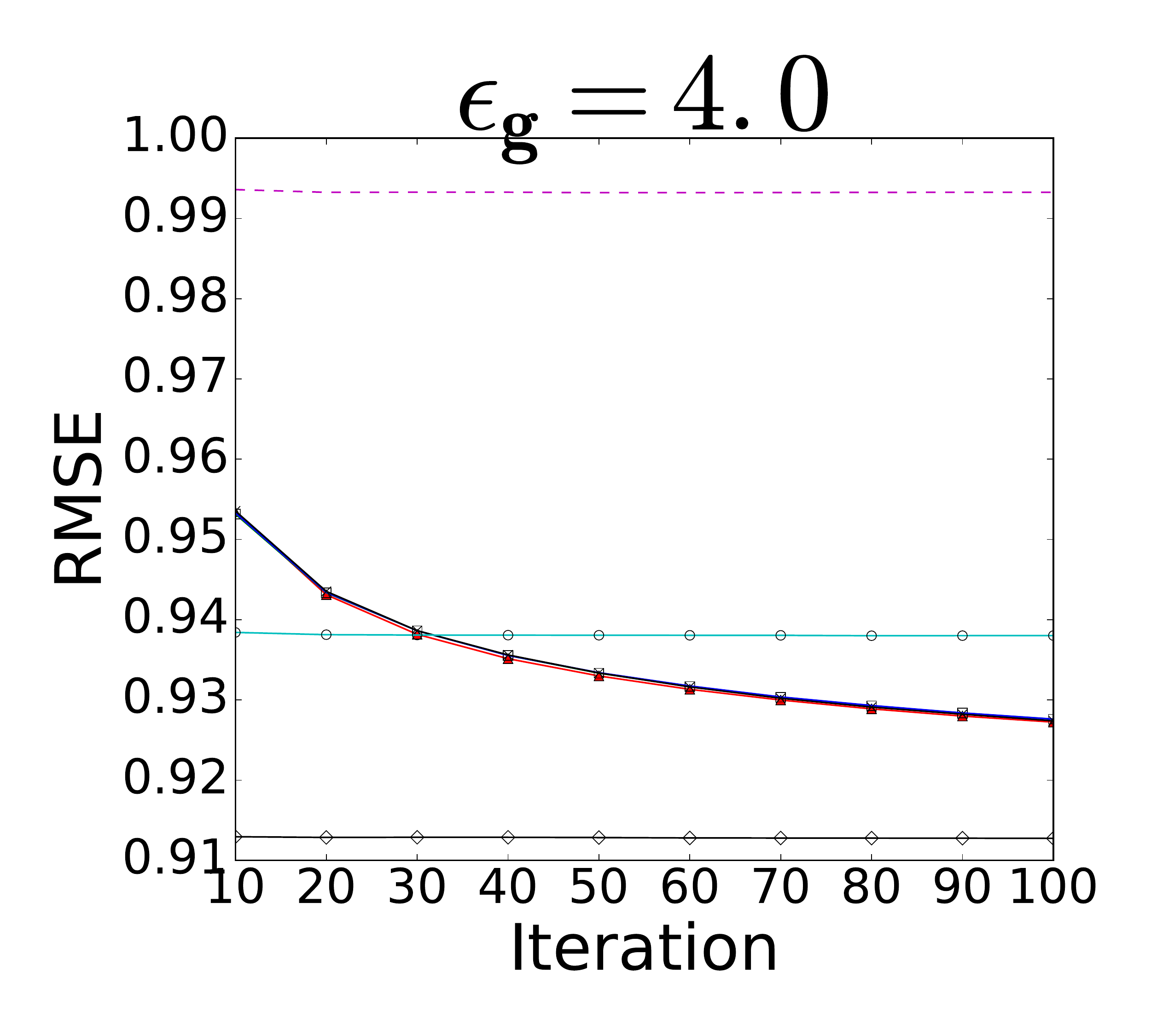}
\includegraphics[width=.35\linewidth]{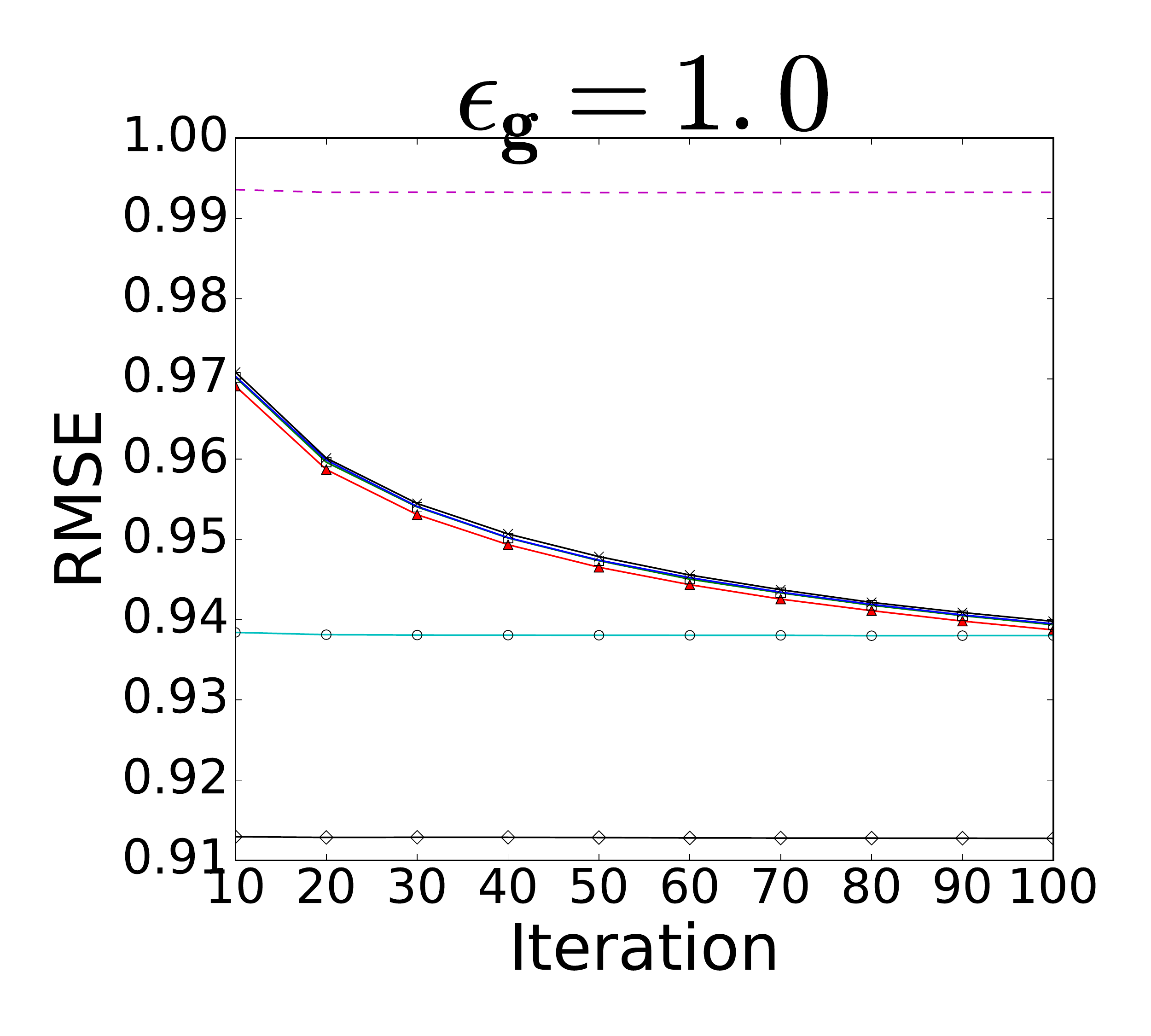}

\includegraphics[width=.35\linewidth]{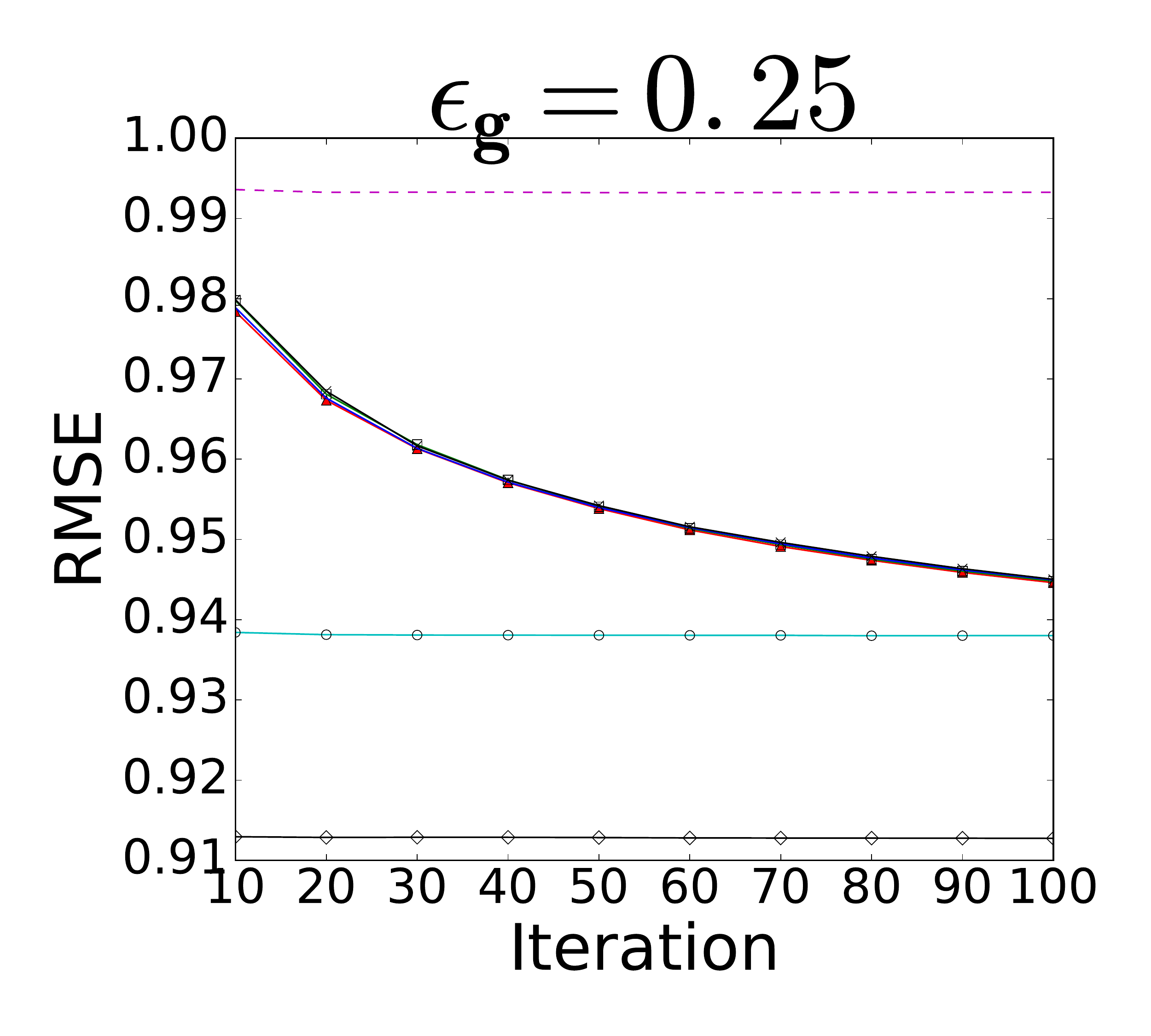}
\includegraphics[width=.35\linewidth]{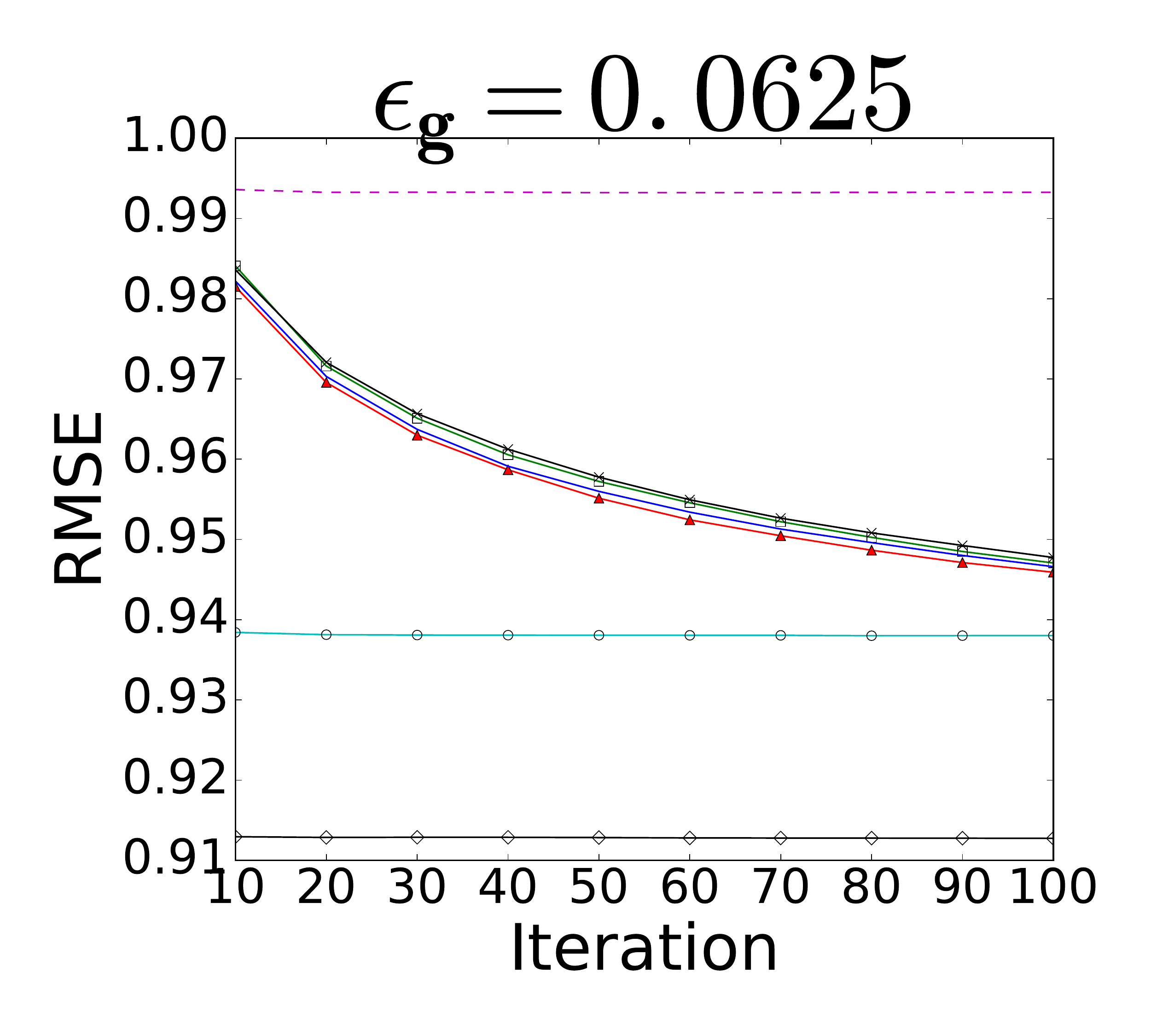}
\includegraphics[width=0.7\linewidth]{legend_epsg.png}
\caption{Task 1: Comparison of different $\epsilon_I$ with fixed $\epsilon_g$.}
\label{fig:ml1M_curve_I}
\end{subfigure}
\caption{MovieLens-1M}
\end{figure}

\begin{figure}
\begin{subfigure}{1.0\linewidth}
\centering
\includegraphics[width=.35\linewidth]{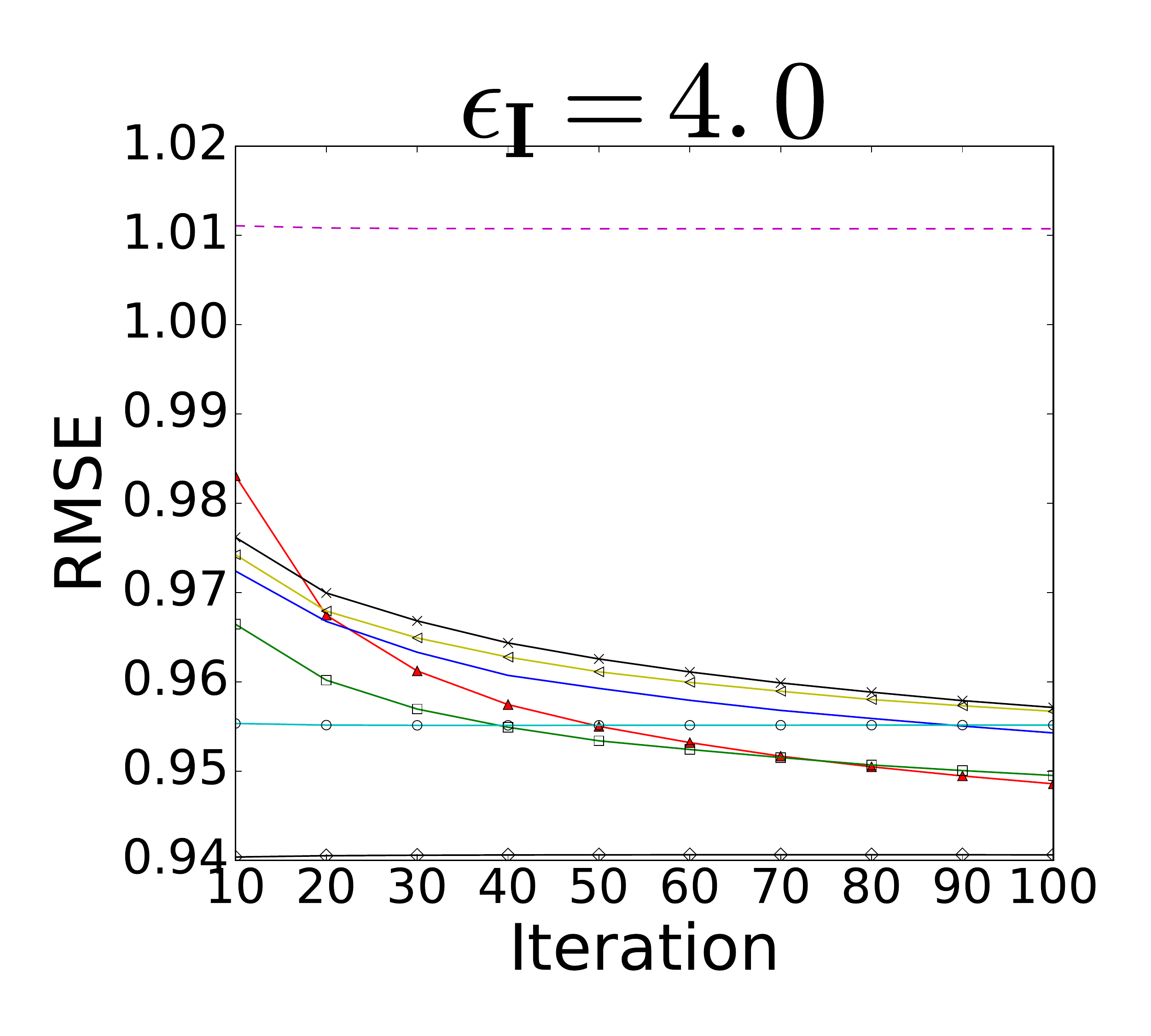}
\includegraphics[width=.35\linewidth]{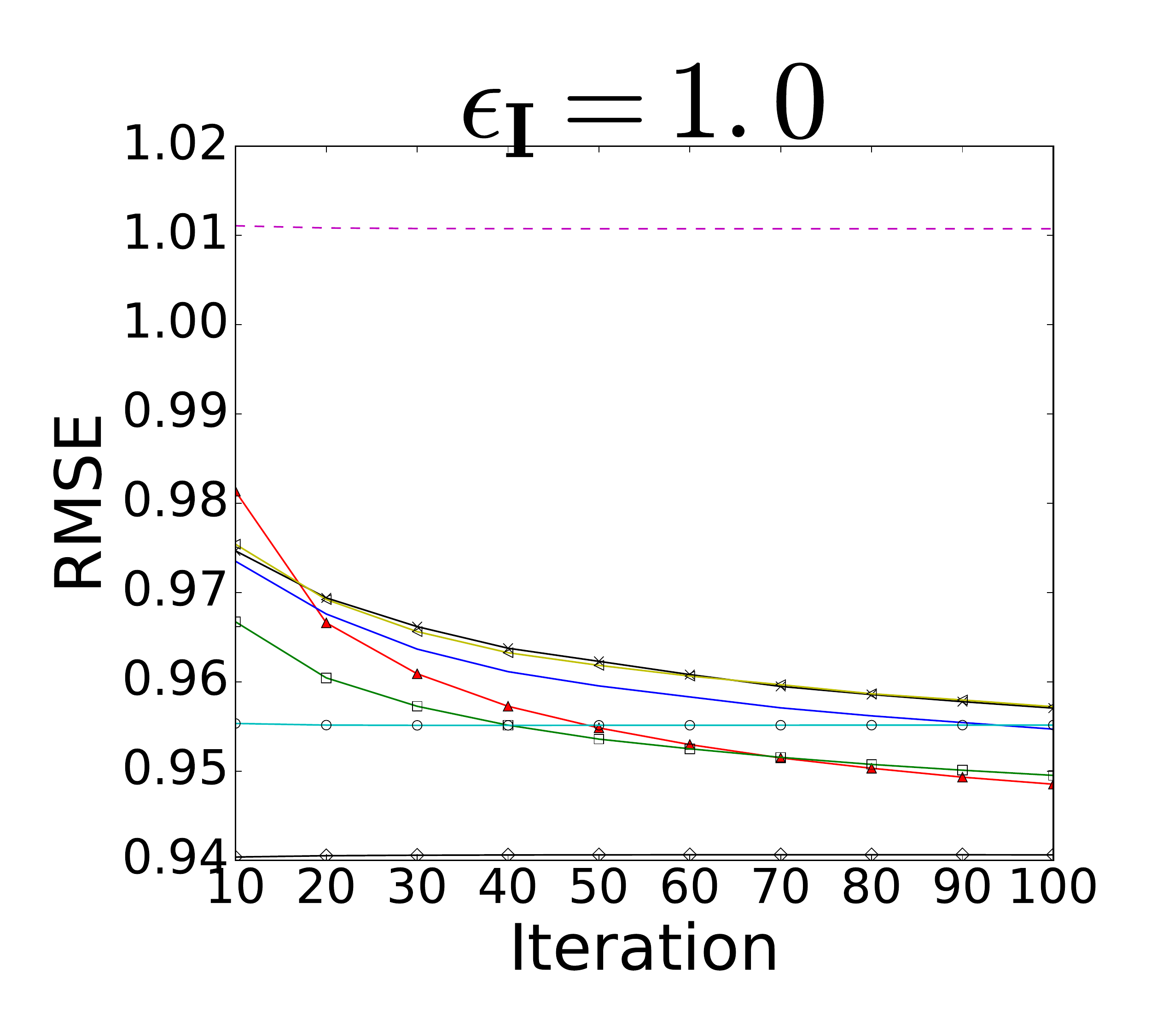}

\includegraphics[width=.35\linewidth]{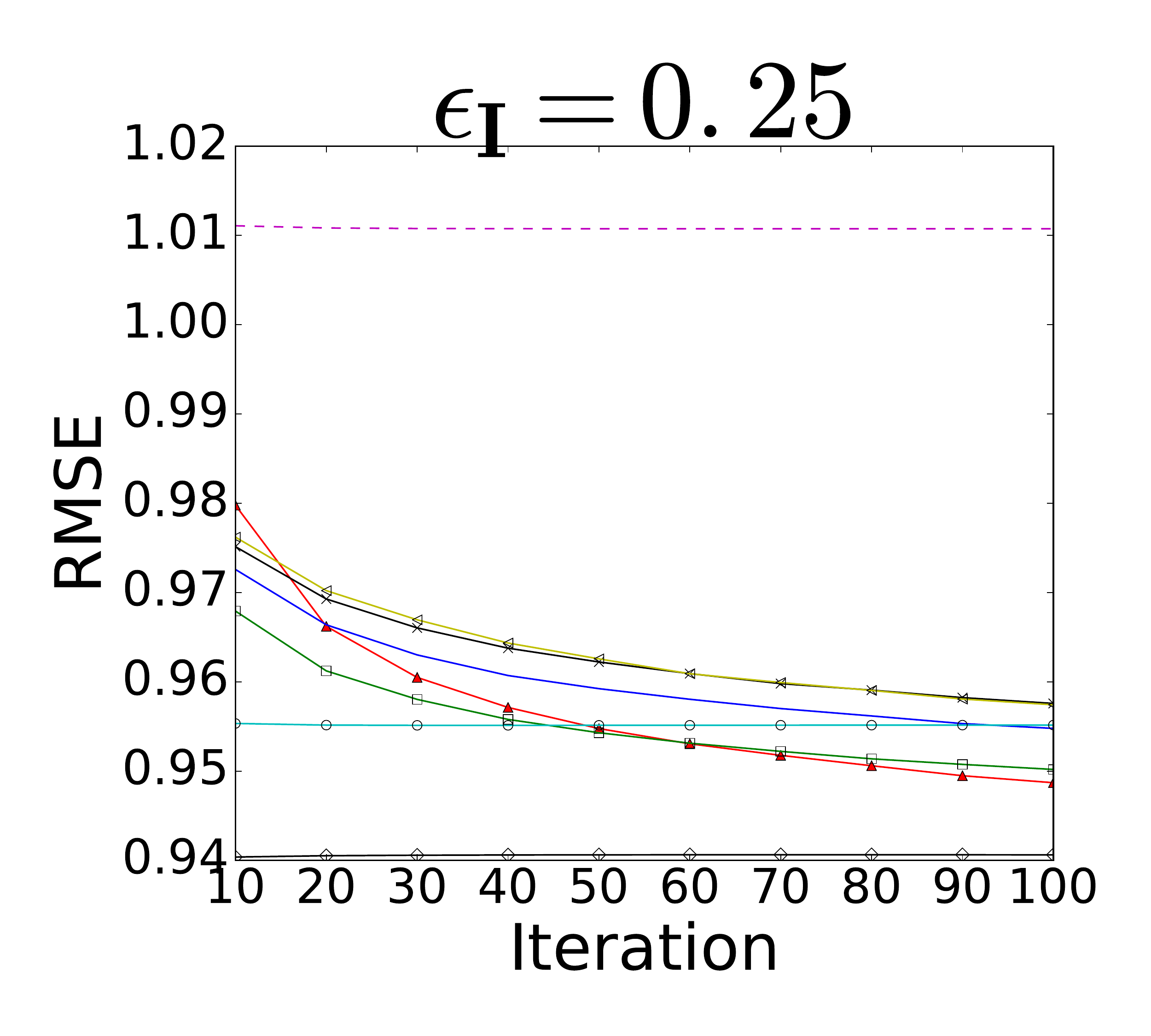}
\includegraphics[width=.35\linewidth]{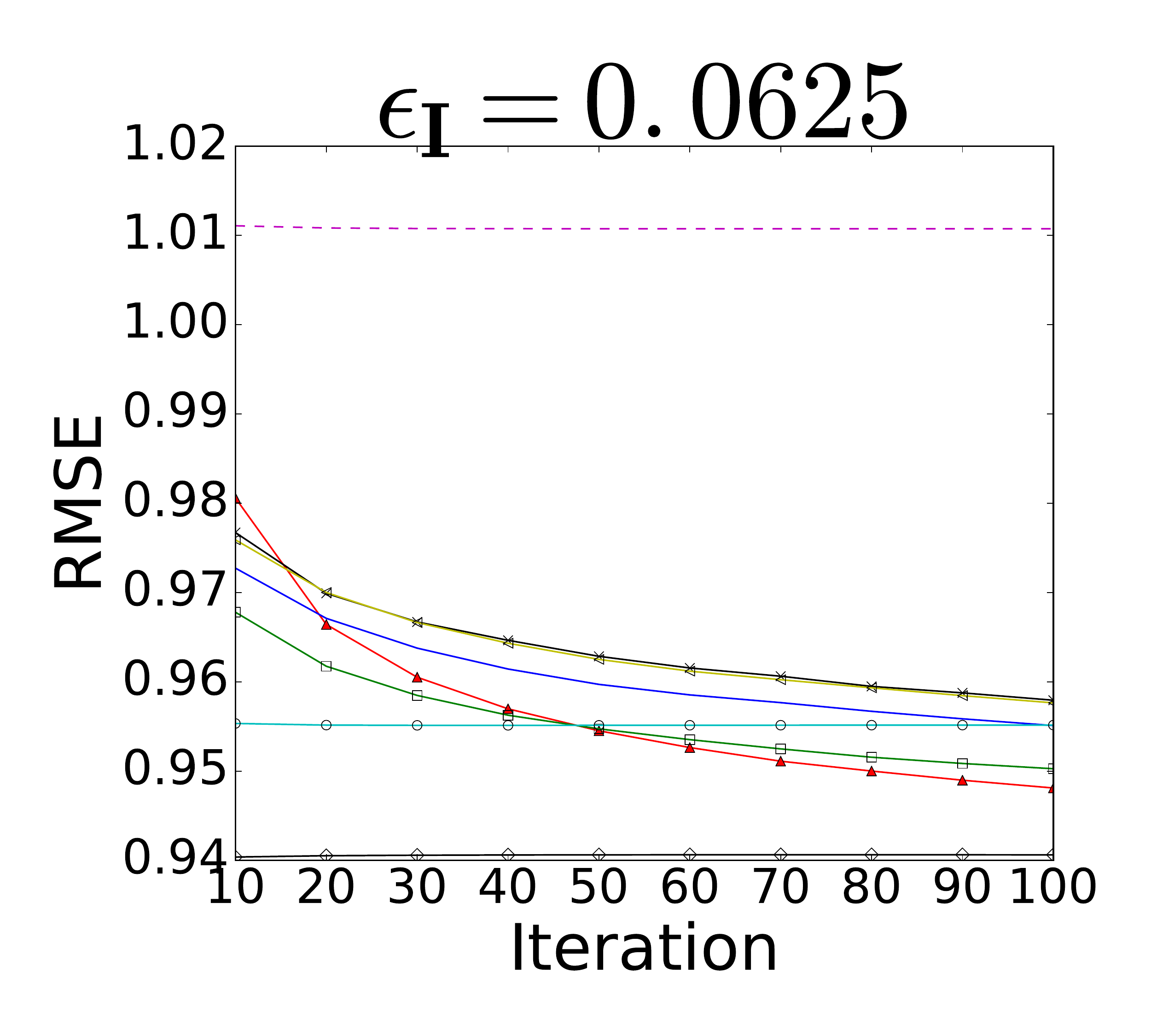}
\includegraphics[width=0.7\linewidth]{legend_epsI.png}
\caption{Task 1: Comparison of different $\epsilon_g$ with fixed $\epsilon_I$.}
\label{fig:Netflix_curve_g}
\end{subfigure}
\begin{subfigure}{1.0\linewidth}
\centering
\includegraphics[width=.35\linewidth]{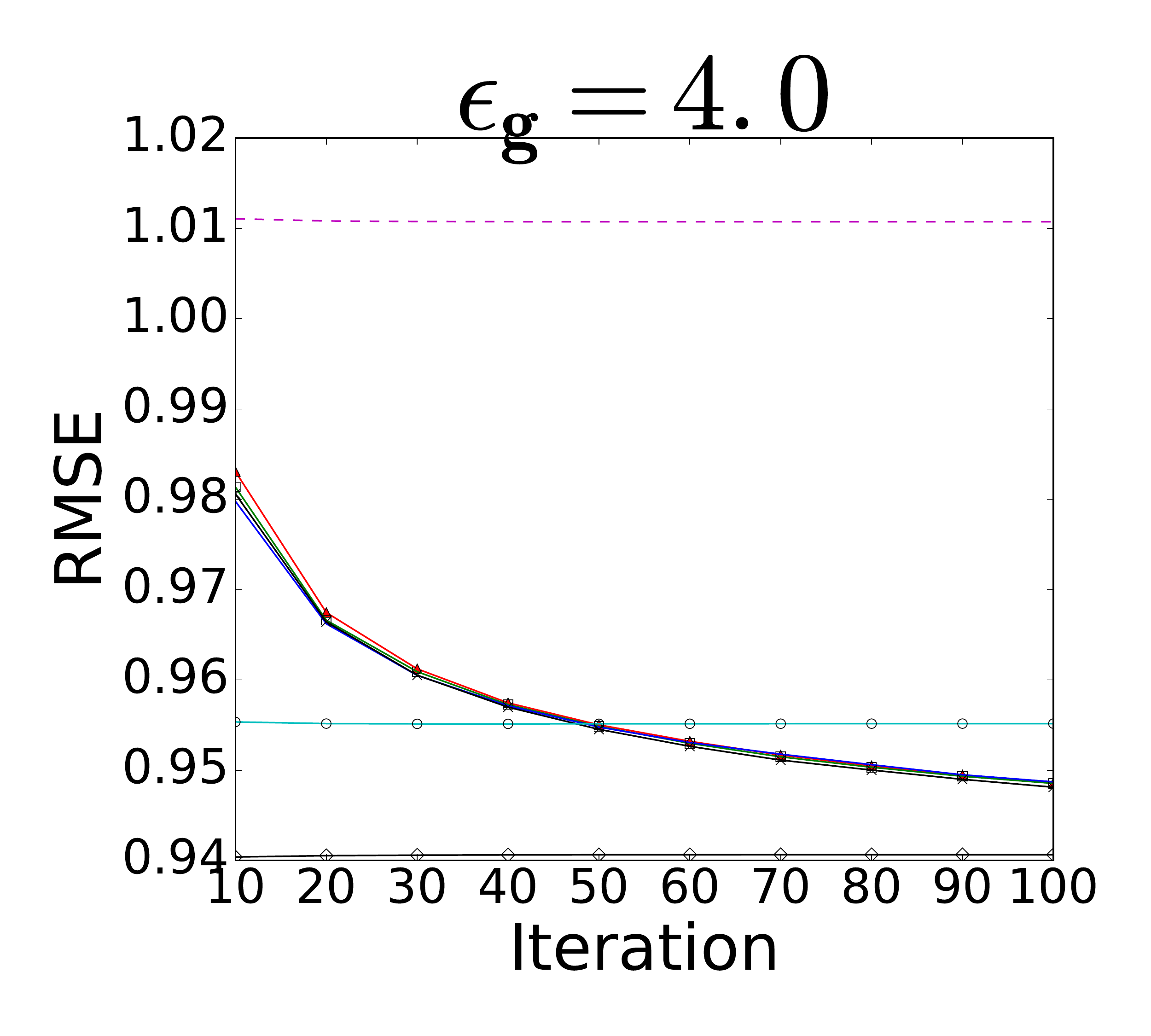}
\includegraphics[width=.35\linewidth]{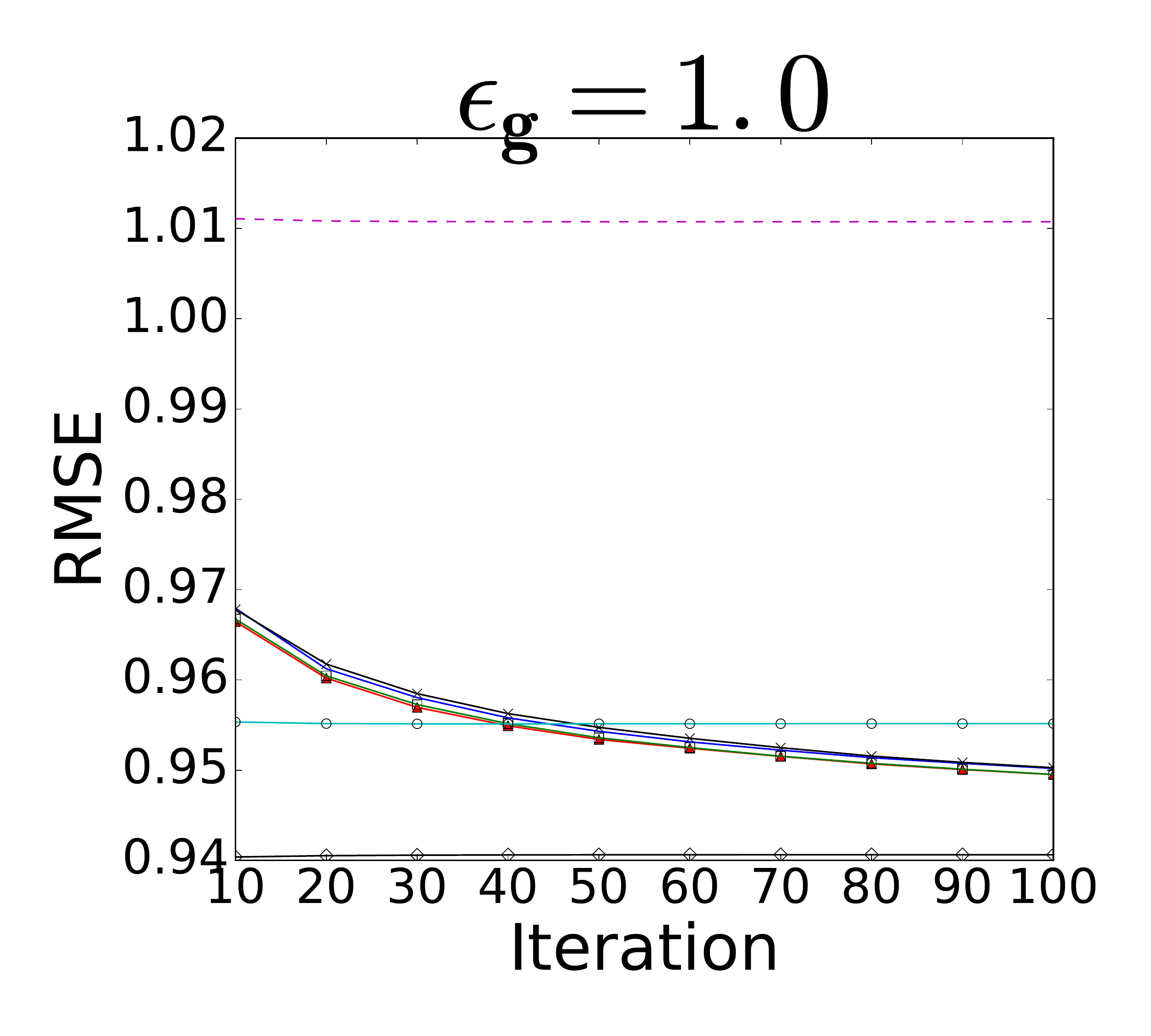}

\includegraphics[width=.35\linewidth]{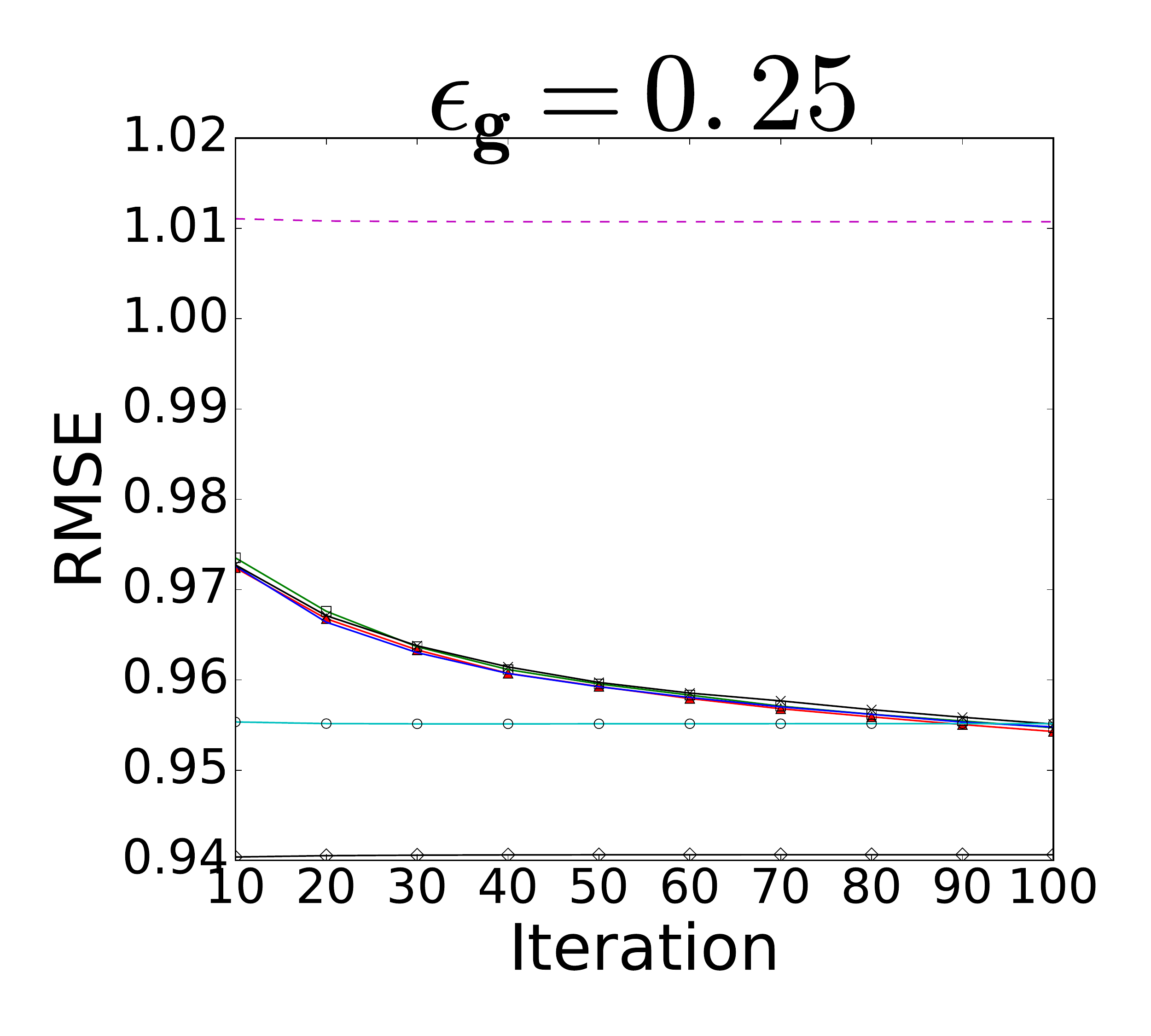}
\includegraphics[width=.35\linewidth]{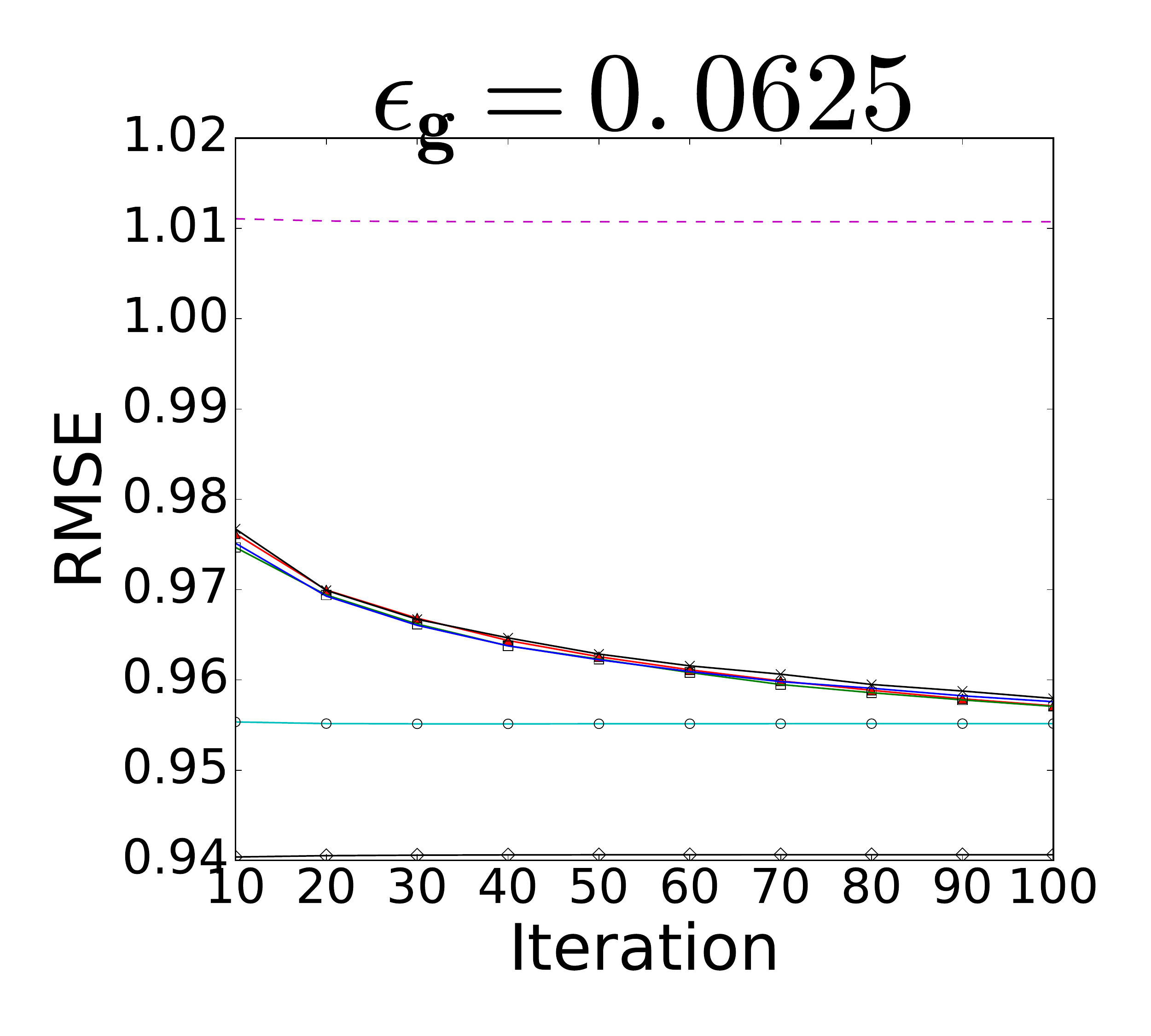}
\includegraphics[width=0.7\linewidth]{legend_epsg.png}
\caption{Task 1: Comparison of different $\epsilon_I$ with fixed $\epsilon_g$.}
\label{fig:Netflix_curve_I}
\end{subfigure}
\caption{Netflix (subsampled)}
\end{figure}

\paragraph{\textbf{Task 1}}
The learning curves on ML-100K are in Fig. \ref{fig:ml100k_curve_g} and \ref{fig:ml100k_curve_I}, the ones on ML-1M are in Fig. \ref{fig:ml1M_curve_g} and \ref{fig:ml1M_curve_I}, and the ones on Netflix are in Fig. \ref{fig:Netflix_curve_g} and \ref{fig:Netflix_curve_I}. 
It can be seen that the distance between ISGLD ($\epsilon=4$) and ISGLD ($\epsilon=2$) is much greater than the ranges that the same curves in SDMF spread. Such results reflect that the performance of SDMF is less sensitive to the privacy budget than ISGLD. This result also shows that SDMF has the advantage to maintain the accuracy while still enjoying the additional protection on \textit{existence} and \textit{model} privacy. While comparing two privacy budgets, in Fig. \ref{fig:ml100k_curve_g}, \ref{fig:ml1M_curve_g}, and \ref{fig:Netflix_curve_g},
we can see $\epsilon_g$ has a negative relation to RMSE, which means more privacy yields lower accuracy on prediction. Although SDMF of $\alpha=\infty$ can already provide comparable performance with ISGLD ($\epsilon=4$), controlling $\alpha$ enjoys arbitrary adjustment of the balance between privacy and accuracy to handle different circumstances.
On the other hand, in Fig. 
\ref{fig:ml100k_curve_I}, \ref{fig:ml1M_curve_I}, and \ref{fig:Netflix_curve_I}, $\epsilon_I$ has less influence on the performance than $\epsilon_g$, which indicates that our two-stage RR algorithm can achieve good privacy without sacrificing too much accuracy. In a nutshell, the RMSE values eventually saturate to reasonable quality as the iteration number increases, and the performance of $\epsilon_g=4$ is indeed comparable to the non-private baseline. Such results implies the proposed SDMF can preserve the rating privacy of users while maintaining the performance of recommender systems.

\begin{figure}
\centering
\includegraphics[width=0.7\linewidth]{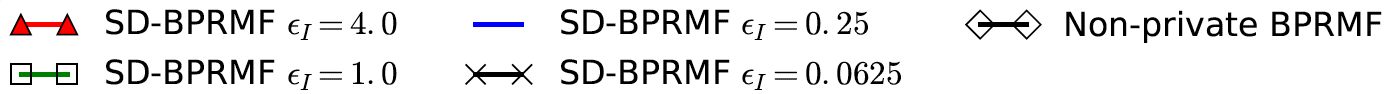}
\begin{subfigure}{0.3\linewidth}
\includegraphics[width=\linewidth]{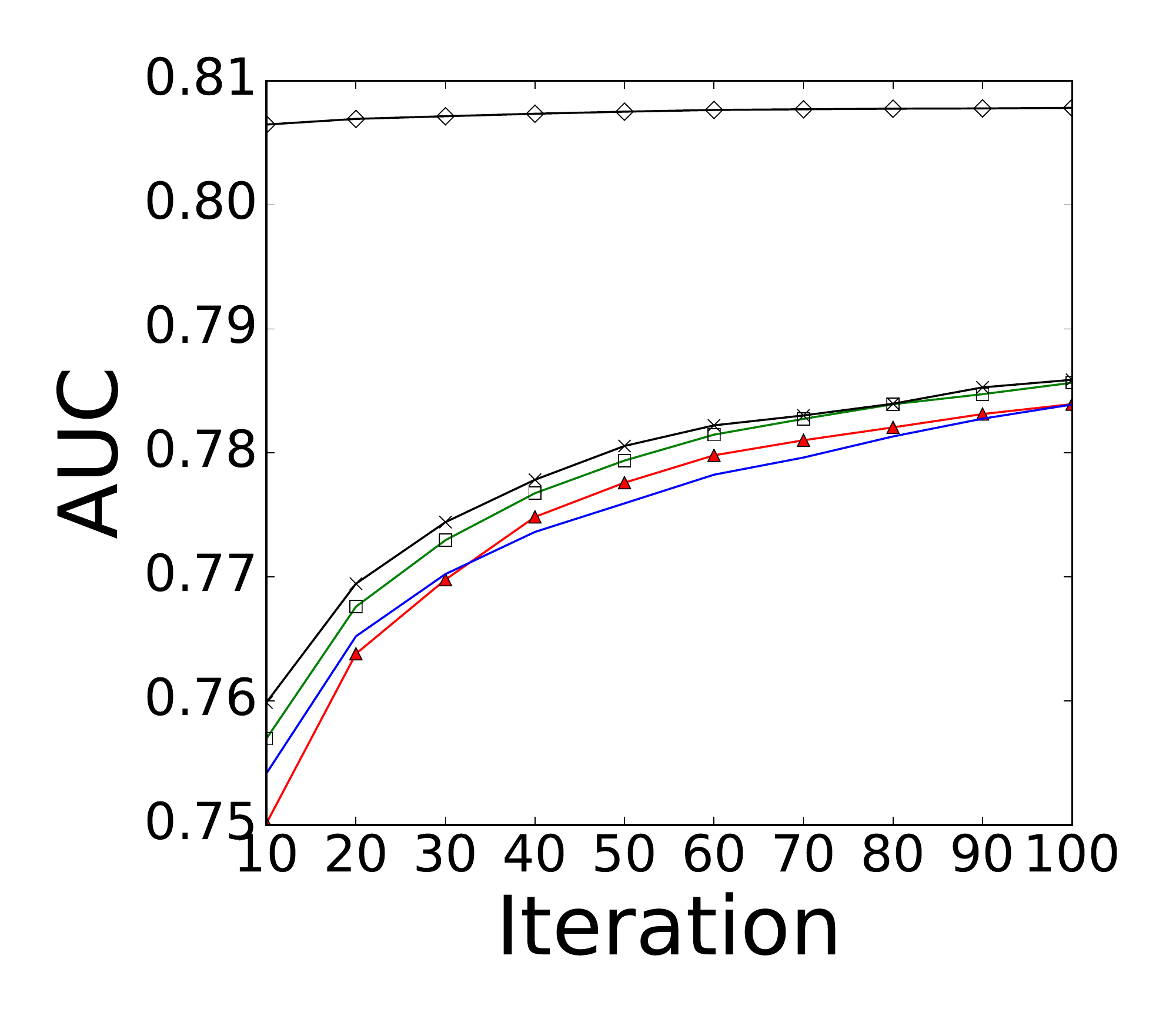}
\caption{ML-100K}
\label{fig:ml100k_bpr_curve}
\end{subfigure}
\begin{subfigure}{0.3\linewidth}
\includegraphics[width=\linewidth]{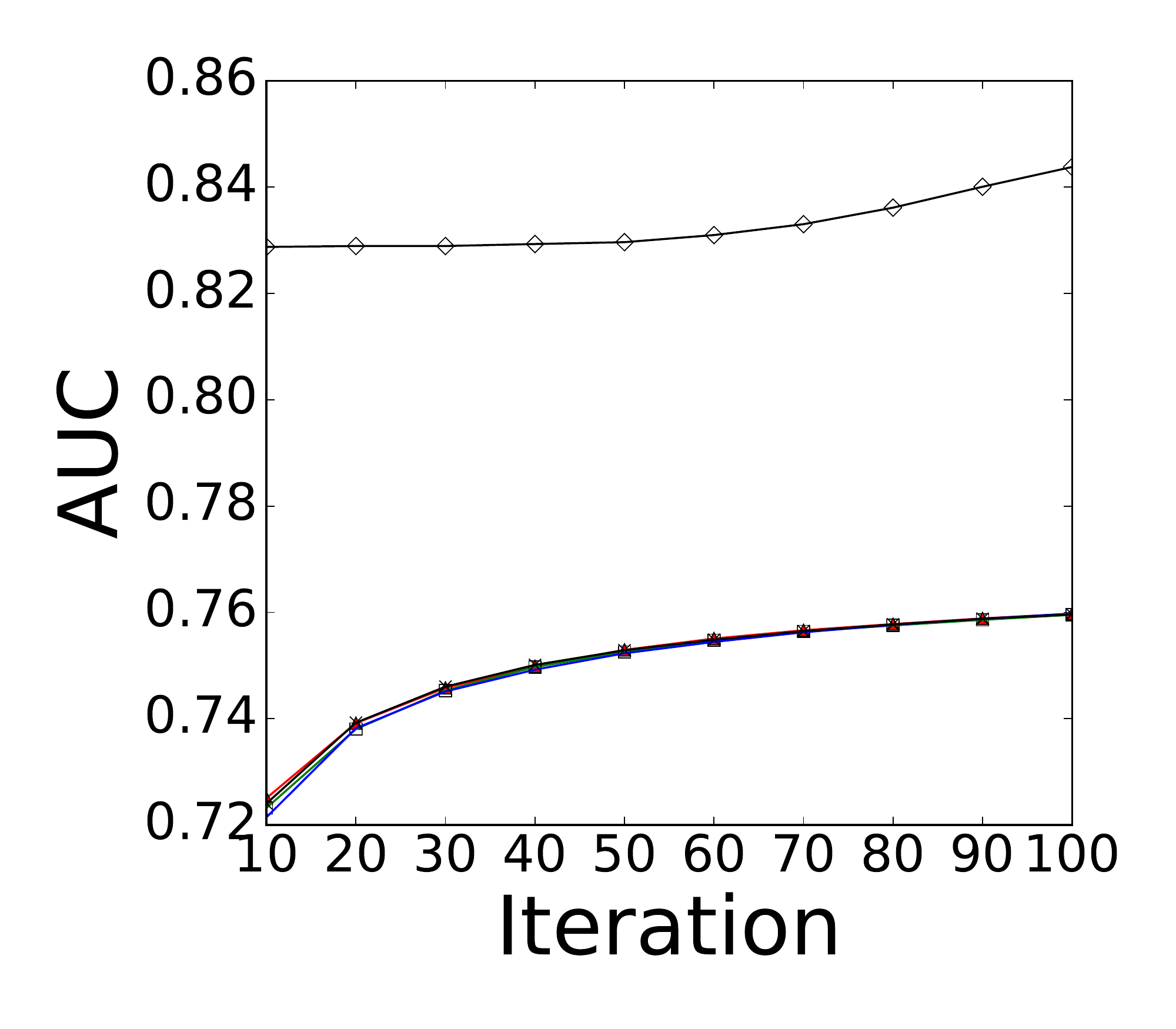}
\caption{ML-1M}
\label{fig:ml1M_bpr_curve}
\end{subfigure}
\begin{subfigure}{0.3\linewidth}
\centering
\includegraphics[width=\linewidth]{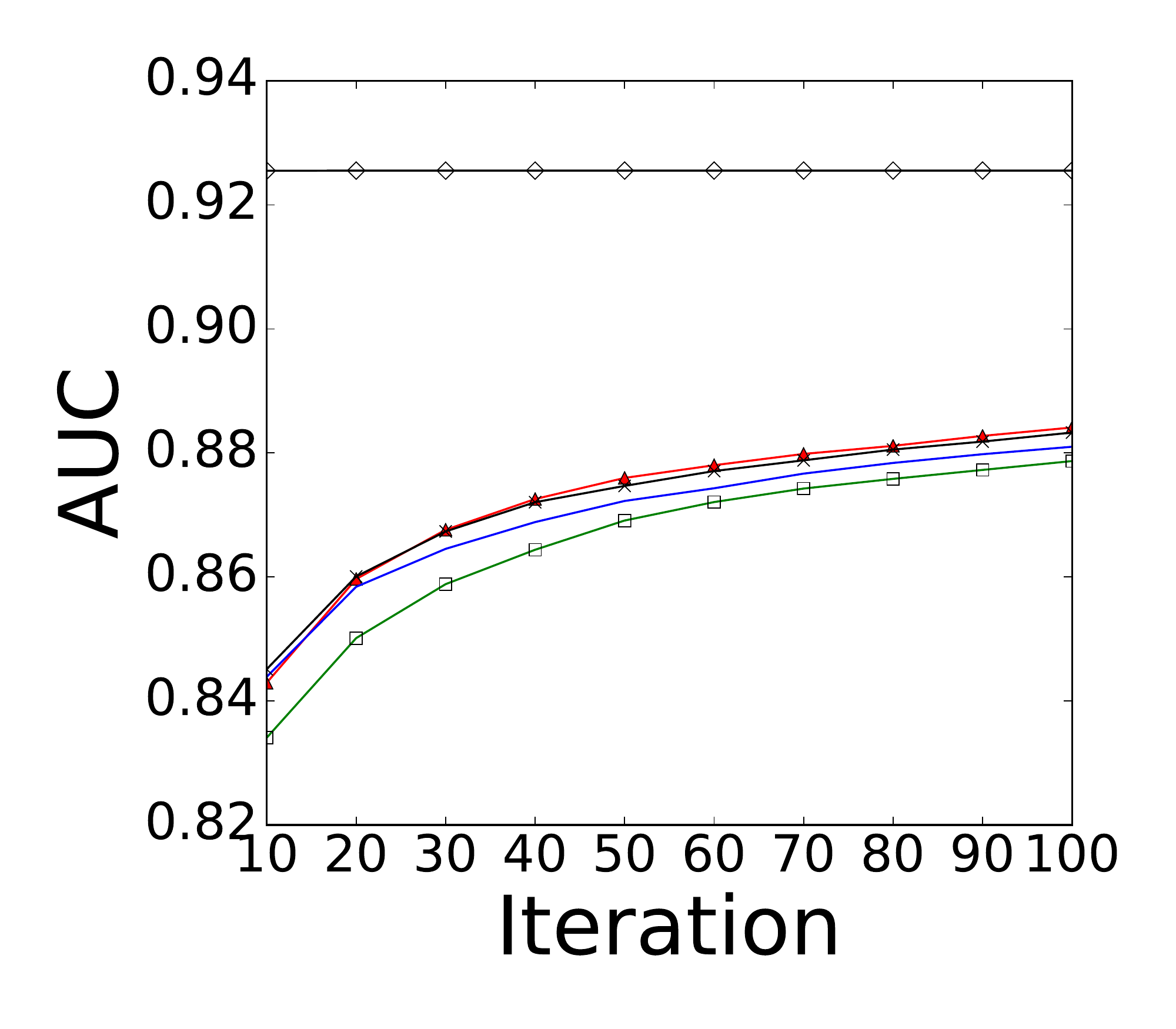}
\caption{Netflix (subsampled)}
\label{fig:Netflix_bpr_curve}
\end{subfigure}

\caption{Task 2: Comparison of different $\epsilon_I$.}
\end{figure}

\paragraph{\textbf{Task 2} }
The results of predicting one-class actions on ML-100K, ML-1M, and Netflix are shown in Fig. \ref{fig:ml100k_bpr_curve}, \ref{fig:ml1M_bpr_curve}, and \ref{fig:Netflix_bpr_curve}, respectively.
Comparing to the non-private baseline, the loss values in AUC on the three datasets are 0.03, 0.07, and 0.05. Similar to Task 1, $\epsilon_I$ has little influence on the performance. However, the gap between private version and non-private version is greater than that in Task 1. In our opinion, this is caused by the loss of information due to PRR (Section \ref{subsec:prr}). In one-class rating action prediction, what PRR does is equivalent to deleting some rating actions from the training set, and thus generates more performance damage than in numerical rating prediction. Nevertheless, as we are the first to exploit differential privacy in one-class recommendation task, such loss is acceptable because the accuracy is still high enough to be with certain utility.

\section{Discussion}
\label{discuss}
Here we discuss how to apply SDCF to Factorization Machines (FM) \cite{rendle2012factorization}, which is another widely-used and robust model for recommendation systems. We choose FM not only because it can be optimized with gradient descent but also because many of CF-based methods such as MF, SVD++, and K-Nearest Neighbors (KNN) can actually be expressed as Factorization Machines with different settings of parameters \cite{rendle2012factorization}.
The basic idea of FM is to learn the interactions between variables. To simply explain FM, let us assume the input data of the prediction task is a matrix $X\in \mathbb{R}^{n\times m}$, where $x\in \mathbb{R}^{m}$ describes a sample with $m$ variables. To model the interactions of $d$-order, for the $c$-th column of total $m$ columns in $X$, FM will learn a bias weight $w_c$ and $d$ latent vectors $v^l_c \in \mathbb{R}^{k_l}$, where $k_l$ is the number of dimensions for the $l$-th latent vector. Therefore, the predicted target (e.g., the predicting scores), denoted by $\hat{y}(x)$, is defined as

\begin{equation*}
\begin{aligned}
\hat{y}(x)=w_0 +\sum_{c=1}^{m}w_cx_c 
+\sum_{l=2}^{d}\sum_{c_1=1}^{m}\dots \sum_{c_d=c_{d-1}+1}^{m}\left(\prod_{\zeta=1}^{l}x_{c_\zeta}\right)\sum_{a=1}^{k_l}\left(\prod_{\zeta=1}^{l}v^l_{c_\zeta a}\right).
\end{aligned}
\end{equation*}

For recommendation systems, each rating will be formulated as an $x$ such that the information about the user and the item can be represented in a format of dummy code as shown in below:

\begin{equation*}
  x=(\underbrace{0,\cdots,0,1,0,\cdots,0}_{|\mathcal{U}|},\underbrace{0,\cdots,0,1,0,\cdots,0}_{|\mathcal{V}|},\underbrace{x_{|\mathcal{U}|+|\mathcal{V}|+1},\cdots,x_{m}}_{\text{other variables}}) ,
\end{equation*}
where other variables describe information such as time, age of user, categories of item, or other attributes about the rating. To incorporate SDCF with FM, we can consider $v^l_c$  and $w_c$ for all $c \leq |\mathcal{U}|$ as personal elements while other bias weights and latent vectors are treated as public elements. In this way, users can download public elements, update their own $v^l_c$ and $w_c$, and then send gradients for other bias weights and latent vectors to the server. 
\section{Conclusions and Future Work}
This paper proposed a framework, SDCF, to preserve privacy of \textit{value}, \textit{model}, and \textit{existence}. The differential privacy of two-stage Randomized Response algorithm and the method we use to compute $gradients$ for unrated items are theoretically justified. 
Experimental results
demonstrate the ability of SDCF to be applied to both prediction tasks of numerical ratings and one-class rating actions without sacrificing too much accuracy.
To improve the feasibility of SDCF for practical usage, in the future we plan to work on the scenario of active learning to lower down the transmission overhead and the scenario of online learning to renew the model without re-training from scratch. Besides, we plan to extend this framework to content-based models and other models that also use gradient descent to learn latent representations (e.g., deep learning framework) so the applications will not be limited to only recommender systems.

\section*{Acknowledgements}
This work was sponsored by Ministry of Science and Technology (MOST) of Taiwan under grants: 106-2628-E-006-005-MY3, 106-2118-M-006-010-MY2, and 106-3114-E-006-002.


\bibliography{mybibfile}

\end{document}